\documentclass{article}

\usepackage{arxiv}

\usepackage[round]{natbib}
\usepackage{booktabs}
\newcommand{\ra}[1]{\renewcommand{\arraystretch}{#1}}

\usepackage[utf8]{inputenc} 
\usepackage[T1]{fontenc}    
\usepackage{hyperref}       
\usepackage{url}            
\usepackage{booktabs}       
\usepackage{amsfonts}       
\usepackage{nicefrac}       
\usepackage{microtype}      
\usepackage{lipsum}
\usepackage{graphicx}
\graphicspath{ {./images/} }

\usepackage{bm}
\usepackage{amsmath}
\usepackage{dsfont}  
\newcommand{\transpose}[1]{#1^\mathsf{T}}
\newcommand{\argmin}{\operatornamewithlimits{arg\,min}}
\DeclareMathOperator{\Corr}{Corr}
\DeclareMathOperator{\Cov}{Cov}
\DeclareMathOperator{\E}{E}

\title{Gaussian Process Models with Low-Rank Correlation Matrices for Both Continuous and Categorical Inputs}

\author{
 Dominik Kirchhoff \\
  Dortmund University of Applied Sciences and Arts\\ Emil-Figge-Str. 42\\
  44227 Dortmund, Germany \\
  \texttt{dominik.kirchhoff@fh-dortmund.de} \\
   \And
 Sonja Kuhnt \\
  Dortmund University of Applied Sciences and Arts\\ Emil-Figge-Str. 42\\
  44227 Dortmund, Germany \\
  \texttt{sonja.kuhnt@fh-dortmund.de} \\
}

\begin{document}
\maketitle
\begin{abstract}
We introduce a method that uses low-rank approximations of cross-cor\-re\-la\-tion matrices in mixed continuous and categorical Gaussian Process models. This new method -- called Low-Rank Correlation (LRC) -- offers the ability to flexibly adapt the number of parameters to the problem at hand by choosing an appropriate rank of the approximation. Furthermore, we present a systematic approach of defining test functions that can be used for assessing the accuracy of models or optimization methods that are concerned with both continuous and categorical inputs. We compare LRC to existing approaches of modeling the cross-cor\-re\-la\-tion matrix. It turns out that the new approach performs well in terms of estimation of cross-correlations and response surface prediction. Therefore, LRC is a flexible and useful addition to existing methods, especially for increasing numbers of combinations of levels of the categorical inputs.
\end{abstract}


\section{Introduction}
Gaussian Process (GP) models are widely used so-called metamodels or surrogate models for replacing expensive black-box function calls with cheap predictions. Such a black-box function can be any function that takes a certain set of inputs and returns one or more continuous quality measures. Examples include elaborate simulation models of real processes that cannot or should not be considered directly, and the assessment of the accuracy of a machine learning method using (possibly nested) cross-validation. Instead of just a point estimate in an unseen point, GP models provide a complete predictive distribution in that point (see, e.g., \citealp{santner}). This enables an iterative search for points in which for instance the expected improvement regarding the current best point is maximized, the famous Efficient Global Optimization (EGO) algorithm \citep{jones}. These points can in turn be evaluated with the black-box function in order to improve the model and find the next interesting point until a certain stop criterion is met.

Original GP models, however, are not directly applicable if one of the inputs is categorical. 
Since many applications have a mixture of continuous and categorical variables, extensions of GP models for the case of mixed inputs are strongly required. Some approaches, which mainly base on incorporating a so-called cross-cor\-re\-la\-tion matrix of the categorial inputs into the covariance function of the GP, have already been introduced (see, e.g., \citealp{joseph}, \citealp{mcmillan}, and \citealp{zhou}). These methods often consist of parameterizations that map a set of box-constrained parameters to a valid correlation matrix in order to reduce the effort of estimating the cross-cor\-re\-la\-tion matrix. The known parameterizations, however, either have a number of parameters that grows quadratically with the number of combinations of levels of the categorical inputs, or they are parsimonious but lack the ability to estimate negative cross-cor\-re\-la\-tions. Note that there are also approaches for group correlation structures (\citealp{qian}, \citealp{roustant}). These work by considering groups of variables rather than individual variables and making assumptions about the correlation structure within groups as well as between different groups. We do not consider these types of parameterizations here as they require knowledge about the variables.

\cite{bouveyron} introduce parsimonious GP models that are able to deal with both numeric and categorical input variables for classification and clustering. Their definition of parsimony, however, is based on the eigen-decomposition of the covariance matrix rather than the number of parameters of the model. Also, we focus on regression.

In this paper, we introduce a new parameterization that offers the ability to choose a feasible number of parameters from a set of possible candidates. The number of parameters depends not only on the number of level combinations of the categorical inputs but also on the rank of the resulting cross-cor\-re\-la\-tion matrix, which can be varied. Yet, even with a lower number of parameters, the new method is able to represent both negative and positive cross-cor\-re\-la\-tions.

We give a short introduction on GP models and known extensions in Section \ref{sec:methods}. The new parameterization is presented in Section \ref{sec:lrc}. There, we also discuss the relationships between various methods and compare their numbers of parameters. In Section \ref{sec:testfuns}, we show how test functions for mixed inputs can be obtained using continuous test functions. We conduct a simulation study to compare the different methods in Section \ref{sec:comp}. Finally, the results are discussed and possible future directions are named in Section \ref{sec:disc}.


\section{Gaussian Process Models for Mixed Inputs}
\label{sec:methods}

We start with the case of purely continuous input variables before we focus on mixed continuous and categorical inputs.

Let $\boldsymbol{x}$ be a vector of $q$ continuous input variables.
A \textit{Gaussian Process (GP) model}, also called \textit{Kriging model}, views the corresponding output $y(\boldsymbol{x})$ of a computer experiment as a realization of a stochastic process
\begin{equation}
Y(\boldsymbol{x}) = \transpose{\boldsymbol{f}}(\boldsymbol{x})\boldsymbol{\beta} + Z(\boldsymbol{x}),
\label{eq:UniversalKriging}
\end{equation}
where $\boldsymbol{f} = \transpose{(f_1(\cdot), \dots, f_p(\cdot))}$ is a vector of known regression functions, $\boldsymbol{\beta} = \transpose{(\beta_1, \dots, \beta_p)}$ is a vector of unknown regression coefficients, and $Z(\boldsymbol{x})$ is a stationary Gaussian stochastic process with mean $\E\left(Z(\boldsymbol{x})\right) = 0$ and covariance function
\begin{equation}
\Cov \left( Z(\boldsymbol{x}_1), Z(\boldsymbol{x}_2) \right) = \sigma^2 R(\boldsymbol{x}_1 - \boldsymbol{x}_2),
\label{eq:Kriging_cov}
\end{equation}
where $\sigma^2$ is the unknown variance of $Z(\boldsymbol{x})$ and $R$ is a correlation function of predefined form \citep{santner}.

Often, the regression terms are replaced by a constant term $\mu$. The resulting model is called \textit{Ordinary Kriging}:
\begin{equation}
Y(\boldsymbol{x}) = \mu + Z(\boldsymbol{x}).
\label{eq:SimpleKriging}
\end{equation}
This substitution is usually affordable because modeling the correlation structure of $Z(\boldsymbol{x})$ is very powerful \citep{jones}.

The correlation function $R$ is often chosen to be of the following form, which is known as the \textit{Mat\'{e}rn$\left(\frac{5}{2}\right)$ correlation function}:
\begin{equation}
R_{\text{mat}}(\boldsymbol{h}) = \prod\limits_{i = 1}^q \exp\left(-\sqrt{5}\frac{|h_i|}{\theta_i}\right)\left( \frac{5 |h_i|^2}{3\theta_i^2} + \frac{\sqrt{5} |h_i|}{\theta_i} + 1\right),
\label{eq:matern5/2}
\end{equation}
where $\boldsymbol{h} = \boldsymbol{x}_1 - \boldsymbol{x}_2$ and $\theta_i > 0$, $i=1,\dots, q$, are unknown parameters \citep{santner}.

Let us now consider the case of mixed inputs $\boldsymbol{w} = \transpose{(\transpose{\boldsymbol{x}}, \transpose{\boldsymbol{v}})}$, where $\boldsymbol{v}$ is a vector of $m$ categorical variables with $v_l$ having $m_l$ levels.

A simple way to deal with mixed inputs is to split the data set into $s = \prod\limits_{l=1}^m m_l$ sub-data sets such that the combination of levels of the categorical inputs is constant in each of the sub-data sets. This results in $s$ smaller data sets that can be considered purely continuous since the constant categorical inputs do not contain any information. In this case, regular Kriging models can be fitted to the sub-data sets. Since there is an individual model for each of the $s$ combinations of levels, this approach is called \textit{Individual Kriging}.

Often Kriging models are applied when the black-box function in hand is expensive. Thus it may not be feasible to evaluate the function so many times that each of the sub-data sets has enough points to build an individual model of sufficient quality.

A more sophisticated approach would aim at sharing information between the different combinations of levels of the categorical variables. This can be done by extending the covariance function to the mixed case:

\begin{equation}
\Cov(Z(\boldsymbol{w}_1), Z(\boldsymbol{w}_2)) = \sigma^2 R(\boldsymbol{x}_1 - \boldsymbol{x}_2) \tau(\boldsymbol{v}_1, \boldsymbol{v}_2),
\label{eq:compcov}
\end{equation}
where $\tau(\boldsymbol{v}_1, \boldsymbol{v}_2)$ is the cross-cor\-re\-la\-tion between the combinations of levels in $\boldsymbol{v}_1$ and $\boldsymbol{v}_2$, which does not depend on the continuous variables. We will denote the combination of levels in $\boldsymbol{v}_i$ by a unique number $c_i \in \{1,\dots,s\}$, where $s$ is the number of level combinations. Thereby, we can write $\tau(\boldsymbol{v}_1, \boldsymbol{v}_2) = \tau_{c_1,c_2}$. In order to be valid, the $(s \times s)$ cross-cor\-re\-la\-tion matrix $(\tau_{i,j})$ must be positive definite with unit diagonal entries (PDUDE) and all of its entries must be in $\left[-1, 1\right]$.

Since these constraints cannot be plugged into a standard nonlinear optimization method, typically para\-metrizations on $(\tau_{i,j})$ are used that ensure these properties automatically.


\cite{joseph} use a constant parameter $c$, $0 < c < 1$, to model cross-cor\-re\-la\-tions:
\begin{equation}
\tau_{i, j} = 
\begin{cases}
c, &\mbox{if $i \neq j$}\\
1, & \mbox{if $i = j$}
\end{cases},
\quad i,j \in \{1,\dots,s\}.
\end{equation}
This approach is called the \textit{compound symmetry} form or \textit{exchangeable correlation} function (EC).

\cite{mcmillan} introduce the so-called \textit{Multiplicative Correlation} structure (MC):
\begin{equation}
\tau_{i, j} = \exp\left(-(\phi_i + \phi_j) \mathds{1}(i \neq j)\right),
\end{equation}
with unknown parameters $\phi_i, \phi_j > 0$ for all $i,j \in \{1, \dots, s\}$, and $\mathds{1}$ is the indicator function.

The most sophisticated approach is by \cite{zhou}, who consider the Cholesky decomposition of the $(s \times s)$ cross-cor\-re\-la\-tion matrix of all level combinations, $\boldsymbol{P} = (\tau_{i, j})$, which is given by
\[
\boldsymbol{P} = \boldsymbol{L}\transpose{\boldsymbol{L}},
\]
where $\boldsymbol{L}$ is a lower triangular matrix with strictly positive diagonal elements. The nonzero part of each row vector of $\boldsymbol{L}$, $(l_{i,1}, \dots, l_{i,i})$, is then modeled as a point on the $i$-dimensional unit hypersphere: For $i = 1$, let $l_{1,1} = 1$ and for $i = 2, \dots, s$, the spherical coordinate system is defined as follows:
\begin{equation}\label{eq:UC}
l_{i,j} =
\left\{
\begin{array}{ll}
\cos\left(\theta_{i,1}\right) &\text{for }j = 1\\
\cos\left(\theta_{i,j}\right) \prod\limits_{k=1}^{j-1} \sin\left(\theta_{i,k}\right) &\text{for }2\le j < i\\
\prod\limits_{k=1}^{j-1} \sin\left(\theta_{i,k}\right)&\text{for }i = j
\end{array}
\right.
\end{equation}
where $\theta_{i, j} \in (0, \pi)$ and $t \in \{2,\dots,r-1\}$. This approach is called the \textit{hypersphere decomposition-based Unrestrictive Correlation} function (UC).

In case of more than one categorical variable, all approaches for modeling the cross-cor\-re\-la\-tion matrix can also be applied to each of these variables rather than the combination of them. This results in $m$ cross-cor\-re\-la\-tion matrices $\tau^l$ of size $(m_l \times m_l)$ $(l=1,\dots,m)$. The cross-cor\-re\-la\-tion matrix of all variables is then obtained by multiplying the relevant individual cross-cor\-re\-la\-tions:
\begin{equation}
(\tau_{c_i,c_j}) = \prod\limits_{l=1}^m \left(\tau^l_{c^l_i,c^l_j}\right),
\label{eq:productform}
\end{equation}
where $c^l_i \in \{1,\dots,m_l\}$ is the level of the $l$-th categorical variable.

Doing this can heavily reduce the number of parameters at the cost of losing interactions between the variables.

Therefore, we will not use the so-called product form \eqref{eq:productform} here. Then, it is possible to transform the $m$ categorical inputs with $s$ combinations of levels into a single categorical variable with $s$ levels $1, \dots, s$. For the sake of simplicity, we assume that this transformation has taken place.


There are a number of different methods for the estimation of the vector $\boldsymbol{\psi}$ that contains all parameters of the correlation functions of the continuous and categorical inputs. Here, we focus on the maximum likelihood (ML) estimation. For other methods, see \cite{santner}.

Up to an additive constant, the log-likelihood is
\begin{equation}
\ell(\boldsymbol{\beta}, \sigma^2, \boldsymbol{\psi}) = 
\frac{1}{2}\left[\vphantom{\frac{1}{\sigma^2}}n \log \sigma^2 + \log(\det(\boldsymbol{R}))
\frac{1}{\sigma^2}\transpose{(\boldsymbol{Y}^n - \boldsymbol{F\beta})}\boldsymbol{R}^{-1}(\boldsymbol{Y}^n - \boldsymbol{F\beta})\right],
\label{likelihood}
\end{equation}
where $\boldsymbol{F} = \transpose{(\boldsymbol{f}(\boldsymbol{w}_1), \dots, \boldsymbol{f}(\boldsymbol{w}_n))}$ is an $(n \times p)$ matrix, $\boldsymbol{R}$ is the $(n \times n)$ correlation matrix of the whole model, i.e., $\boldsymbol{R}_{i,j} = \Corr \left(Z(\boldsymbol{w}_i), Z(\boldsymbol{w}_j)\right)$ $= R(\boldsymbol{x}_i - \boldsymbol{x}_j)\cdot \tau(\boldsymbol{v}_i, \boldsymbol{v}_j)$, and $\boldsymbol{Y}^n$ is the response vector.

Given $\boldsymbol{\psi}$, the ML estimate of $\boldsymbol{\beta}$ is the generalized least squares estimate
\begin{equation}
\widehat{\boldsymbol{\beta}} = \left(\transpose{\boldsymbol{F}}\boldsymbol{R}^{-1}\boldsymbol{F}\right)^{-1} \transpose{\boldsymbol{F}}\boldsymbol{R}^{-1} \boldsymbol{Y}^n
\label{MLbeta}
\end{equation}
and the ML estimate of $\sigma^2$ is
\begin{equation}
\widehat{\sigma}^2 = \frac{1}{n}\transpose{\left(\boldsymbol{Y}^n - \boldsymbol{F\widehat{\beta}}\right)}\boldsymbol{R}^{-1}\left(\boldsymbol{Y}^n - \boldsymbol{F\widehat{\beta}}\right).
\label{MLsigma}
\end{equation}
Plugging Equations \eqref{MLbeta} and \eqref{MLsigma} in the log-likelihood \eqref{likelihood} yields
\begin{equation}
\ell(\widehat{\boldsymbol{\beta}}, \widehat{\sigma}^2, \boldsymbol{\psi}) = -\frac{1}{2}\left[n \log \widehat{\sigma}^2 + \log(\det(\boldsymbol{R})) + n\right],
\label{likelihood2}
\end{equation}
which only depends on $\boldsymbol{\psi}$. The ML estimate chooses $\widehat{\boldsymbol{\psi}}$ to maximize \eqref{likelihood2}, and can be written equivalently as
\begin{equation}
\widehat{\boldsymbol{\psi}} = \argmin_{\boldsymbol{\psi}}n \log \widehat{\sigma}^2 +\log(\det{\boldsymbol{R}}).
\end{equation}

Given the estimated parameters, the \textit{empirical best linear unbiased predictor} (EBLUP) of $\boldsymbol{Y}^n$ at any input $\boldsymbol{w}_0$ is
\begin{equation}
\widehat{Y}(\boldsymbol{w}_0) = \transpose{\boldsymbol{f}}\widehat{\boldsymbol{\beta}}  + \transpose{\widehat{\boldsymbol{r}}_0} \widehat{\boldsymbol{R}}^{-1} \left( \boldsymbol{Y}^n - \boldsymbol{F}\widehat{\boldsymbol{\beta}}\right),
\label{EBLUPUK}
\end{equation}
where $\widehat{\boldsymbol{r}}_0 = \transpose{\left(\Corr(\boldsymbol{w}_0, \boldsymbol{w}_1), \dots, \Corr(\boldsymbol{w}_0, \boldsymbol{w}_n)\right)}$ and $\widehat{\boldsymbol{R}}$ is the estimated $(n \times n)$ correlation matrix.

This ML estimation of course requires a suitable data set which typically is obtained by applying a space-filling design. For mixed inputs, there are a number of different space-filling designs. The most straightforward way would be to use the same Latin Hypercube Design (LHD) for all the levels of the categorical variable. Then, we would not use the possibility to vary the continuous variables for different levels of the categorical variable. However, sharing as much information across the levels as possible could be beneficial for the model accuracy. Unrelated LHDs for each of the levels, on the other hand, would hamper the estimation of the cross-cor\-re\-la\-tions. Motivated by this trade-off, \cite{huang} develop the class of \textit{Clustered Sliced Latin Hypercube Designs (CSLHDs)}. The complete CSLHD is an LHD, all its slices (i.e., the separate designs for the levels of the categorical variable) form LHDs, and for each point in any slice of the design there is another point in a different slice that is nearby (thus the term ``clustered'').

\section{Low-Rank Correlation}
\label{sec:lrc}

In this section, we look at an alternative approach to model the cross-cor\-re\-la\-tion matrix for a GP model with both continuous and categorical inputs. The aim is to get a parameterization that is both parsimonious and flexible. In Section \ref{sec:lrcdef}, we present this parameterization. In Section \ref{sec:numpars}, we consider the relationships between the different methods and compare their numbers of parameters.

\subsection{Definition}
\label{sec:lrcdef}

One way to obtain a symmetric and psd $(s \times s)$ matrix is to multiply an arbitrary real $(s \times r)$ matrix $\bm{Q}$ with its transpose. Here, $r$ can be chosen arbitrarily. This approach can be used to model the cross-cor\-re\-la\-tion matrix, where the elements of $\bm{Q}$ have to be parameterized such that the diagonal elements of $\bm{Q} \transpose{\bm{Q}}$ are equal to 1 and all other elements are in $\left[-1,1\right]$. Here, we consider the case of $r$ being as small as possible. Since the rank of $\bm{Q} \transpose{\bm{Q}}$ is equal to the rank of $\bm{Q}$, we have chosen the name \textit{Low-Rank Correlation} (LRC). For $r = 1$, $\bm{Q} \transpose{\bm{Q}}$ has unit diagonal entries if and only if $\bm{Q} = \transpose{(1, \dots, 1)}$, but then $\bm{Q} \transpose{\bm{Q}}$ is the all-ones matrix. In the following definition, we therefore focus on the case of $r = 2$ and use a simple parameterization to generate $\bm{Q}$.

We denote the columns of $\bm{Q}$ by $\bm{l}_1$ and $\bm{l}_2$, which are real column vectors of length $s$.
\begin{align}
\bm{P} = \bm{Q} \transpose{\bm{Q}} = (\bm{l}_1, \bm{l}_2) \transpose{(\bm{l}_1, \bm{l}_2)}
\end{align}
is symmetric and psd. Since $\bm{P}$ must have unit diagonal entries, we get the condition
\[
l_{1i}^2 + l_{2i}^2 = 1\quad \forall i \in \{1, \dots, s\},
\]
which is obviously fulfilled for $l_{1i} = \sin\left(\theta_i\right)$ and $l_{2i} = \cos\left(\theta_i\right)$ for any $\bm{\theta} \in \left[0, 2\pi\right]^s$.

Also, the entries $\tau_{i,j}$ of $\bm{P}$ simplify to
\begin{equation}
\begin{split}
\tau_{i,j} &= \sin\left(\theta_i\right) \sin\left(\theta_j\right) + \cos\left(\theta_i\right) \cos\left(\theta_j\right)\\
&= \cos\left(\theta_i - \theta_j\right),
\end{split}
\label{eq:lrctau}
\end{equation}
which means that
\[
\tau_{i,j} \in \left[-1, 1\right] \quad \forall i, j \in \{1,\dots,s\}.
\]
In order to ensure the invertibility of $\bm{P}$, we add a small nugget to the diagonal elements, which makes any psd matrix pd. We then have to rescale the whole matrix such that $\tau_{i,i} = 1$.
$\bm{P}$ is a valid correlation matrix with the parameterization defined above.

\cite{rapisarda} derive a very similar parameterization from a geometrical point of view, motivated by applications in finance. They introduce a rank-$r$ decomposition of a correlation matrix $\boldsymbol{C}$ for a general rank $2 \le r < s$, i.e.,
\begin{equation}
\boldsymbol{C} \simeq Q \transpose{Q},
\end{equation}
where $Q = (q_{i,j})$ is an $(s \times r)$ matrix with $q_{1,1} = 1$ and
\begin{equation}\label{eq:LRC}
q_{i,j} = \left\{
\begin{array}{ll}
\cos\left(\theta_{i,1}\right) &\text{for }j = 1\\
\cos\left(\theta_{i,j}\right) \prod\limits_{k=1}^{j-1} \sin\left(\theta_{i,k}\right)&\text{for }2 \le j < \min(i, r)\\
\prod\limits_{k=1}^{j-1} \sin\left(\theta_{i,k}\right)&\text{for }j = \min(i, r)\\
0 &\text{for }\min(i, r) < j \le s.
\end{array}
\right.
\end{equation}
Note that now the first row of the matrix $Q$ always is $\transpose{(1, 0)}$, thus saving one parameter. In fact, when looking at Equation \eqref{eq:lrctau} it is obvious that only the differences between the parameters have an impact on the cross-cor\-re\-la\-tion matrix rather than their actual values. We will continue with this definition even for the rank-2 approximation of the cross-cor\-re\-la\-tion matrix.


We will denote this approximation \textit{Low-Rank Correlation} function $\left(\text{LRC}_r\right)$, where the subscript denotes the rank of the approximation. In the next subsection, we consider the relationships betweens the various methods and compare the number of parameters that have to be estimated.

\subsection{LRC in comparison to EC, MC, and UC}
\label{sec:numpars}

When comparing the definitions of LRC and UC (Equations \eqref{eq:LRC} and \eqref{eq:UC}), it is obvious that they are very similar to each other. In fact, LRC is a special case of UC, obtained by setting $\theta_{i,r} = 0$ for each $i > r$. Subsequent parameters $\theta_{i,r+1}, \dots, \theta_{i, i-1}$ then do not influence the resulting correlation matrix because they only appear in products where they are multiplied with $\sin\left(\theta_{i,r}\right) = 0$. Thus, they can be chosen freely.

Moreover, UC is an unrestrictive parameterization with a one-to-one correspondence between a PDUDE matrix and the parameter vector $\bm{\theta}$. That is, an arbitrary PDUDE matrix can be parameterized using a certain $\bm{\theta}$ and each $\bm{\theta}$ results in a  PDUDE matrix \citep{zhou}. Therefore, each matrix genereated by EC and MC (and any other method for generating PDUDEs) can also be obtained with UC using a particular $\bm{\theta}$. However, the relationships are much more complex than setting some of the $\theta$'s to 0 such that EC and MC cannot be viewed as special cases of UC. For instance, consider $s=3$. In order to obtain the matrix given by EC,

\[
\bm{P}=\begin{pmatrix}
1 &  &  \\ 
c & 1 &  \\ 
c & c & 1
\end{pmatrix},
\]
one has to solve the following system of equations,

\begin{align*}
\cos\left(\theta_{2,1}\right) &= c\\
\cos\left(\theta_{3,1}\right) &= c\\
\cos\left(\theta_{2,1}\right)\cos\left(\theta_{3,1}\right)+\sin\left(\theta_{2,1}\right)\sin\left(\theta_{3,1}\right)\cos\left(\theta_{3,2}\right) &= c,
\end{align*}

resulting in $\theta_{2,1} = \cos^{-1}\left(c\right)$, $\theta_{3,1} = \cos^{-1}\left(c\right)$, and $\theta_{3,2} =\cos^{-1}\left(\frac{c}{c+1}\right)$.

MC's matrix
\[
\bm{P}=\begin{pmatrix}
1 &  &  \\ 
\exp\left(-\left(\phi_2 + \phi_1\right)\right) & 1 &  \\ 
\exp\left(-\left(\phi_3 + \phi_1\right)\right)  & \exp\left(-\left(\phi_3 + \phi_2\right)\right)  & 1
\end{pmatrix}
\]
leads likewise to  $\theta_{2,1} = \cos^{-1}\left(\exp\left(-\left(\phi_2 + \phi_1\right)\right)\right)$, $\theta_{3,1} = \cos^{-1}\left(\exp\left(-\left(\phi_3 + \phi_1\right)\right)\right)$, and $\theta_{3,2} =$\\
$
 \cos^{-1}\left(\frac{\exp\left(-\left(\phi_3 + \phi_2\right)\right) - \exp\left(-\left(\phi_2 + \phi_1\right)\right)\exp\left(-\left(\phi_3 + \phi_1\right)\right)}{\sqrt{1-\exp\left(-\left(\phi_2 + \phi_1\right)\right)^2}\sqrt{1-\exp\left(-\left(\phi_3 + \phi_1\right)\right)^2}}\right).
$

Of course, these relationships get even more complicated with growing $s$.
These examples show that EC, MC, and UC are systematically different approaches in the sense that there are no simple mappings between the parameters of EC and MC to UC (and thus LRC, too).

Table \ref{tab:numberpars} contrasts the numbers of parameters that each of the approaches named above uses for modeling the cross-cor\-re\-la\-tions of the categorical input. We can see that the number of parameters of both MC and LRC$_r$ with a fixed rank are linear in the number of levels of the categorical variable whereas UC has a quadratic number of parameters. This can lead to computational problems as soon as $s$ gets bigger, especially in the case of few observations.

\begin{table*}
	\centering
	\ra{1.3}
	\begin{tabular}{@{}ll@{}}
		\toprule
		Model& Number of parameters\\
		\midrule
		EC                   & 1\\
		MC                  & $s$\\
		LRC$_2$			& $s-1$\\
		LRC$_3$			& $2s -3$\\
		LRC$_r$			& $(r-1)(s - \frac{r}{2})$\\
		UC                   & $\dfrac{s^2 - s}{2}$\\
		\bottomrule
	\end{tabular}
	\caption{Numbers of parameters needed for modeling the categorical inputs. Here, $s$ is the number of levels of the categorical variable and $r$ is the rank of the cross-cor\-re\-la\-tion matrix of LRC.}\label{tab:numberpars}
\end{table*}

We will compare the accuracy of the cross-cor\-re\-la\-tion matrix estimation and of the prediction of the different methods with a simulation study in Section \ref{sec:comp}. Before, we need a set of suitable test functions.

\section{Test Functions}\label{sec:testfuns}

In contrast to purely continuous inputs, for mixed inputs there are not many known test functions. One testbed containing many continuous test functions is the Black-Box Optimization Benchmarking (BBOB) set of noiseless test functions of the COCO (``COmparing Continuous Optimizers'') platform \citep{coco}.
The \textsf{R} package \texttt{smoof} \citep{smoof} is an interface to the BBOB set of functions as well as to many different Single- and Multi-Objective Optimization test Functions (SMOOF) that are frequently used in the literature.

 In most papers concerning the design and analysis of computer experiments with mixed inputs, the test functions are often a composition of relatively simple continuous functions for different levels of the categorical variable. \cite{zhou}, e.g., use two test functions composed of three sinoidal and three polynomials of order 2, respectively, where all functions depend on only one continuous variable. They also consider a data set from a data center computer experiment that is used for predicting airflow and heat transfer in the electronic equipment of an air-cooled data center. In this example, there are five quantitative  and three qualitative inputs with 24 combinations of levels. In the last example of their paper, they consider the so-called Borehole function, a function that models the flow of water through a borehole. Since this function has purely continuous inputs, three of the variables were discretized so that only three values for each of these inputs were considered.

In this paper, we generalize their approach with the Borehole function by using an arbitrary single-objective continuous test function with at least two dimensions and then discretize one of its dimensions. Here, we call this procedure ``slicing''. The continuous test function can be sliced in various ways. Here, we slice the dimension to be sliced between its bounds equidistantly, i.e., if $l$ and $u$ are the lower and upper bounds, respectively, the slice positions are $\mathrm{pos}_i = l + (i - 1)\dfrac{u - l}{s - 1}$. This procedure, however, ignores the position of the global optimum. Since it can be crucial to know the exact position and value of the global optimum for benchmarking purposes, one can exchange one of the slice positions with the position of the optimum, thus ensuring that the optimum of the continuous function still exists in the sliced version of it. Although in this paper we do not require the global optima, we nevertheless for each sliced function exchange the slice position that is closest to the global optimum, considering only the dimension to be sliced, with the optimum's position. E.g., assume that the original continuous function had three dimensions, the global optimum was in $\left(x_1^*, x_2^*, x_3^*\right)$, and we wanted to slice the first dimension. Then, we exchange the $i$-th initial slice position $\mathrm{pos}_i$ with $x_1^*$, where $i = \argmin_i \left|\textrm{pos}_i - x_1^*\right|$ . In the case that $i$ is ambiguous, i.e., two values in \texttt{pos} have the same distance to the dimension's position of the global optimum, only the lower of the two corresponding slice positions is swapped.



Here, we consider four different test functions, namely the Ackley, Alpine N. 1, Deflected Corrugated Spring, and Double-Sum functions, which all have unique global minima. All of them are real functions depending on three continuous inputs, where the first dimension is sliced into $s=4$ and $s=6$ slices, respectively. For details on how we implemented the test functions, see Section \ref{sec:slicedSmoof} in the appendix. Table \ref{tab:testfunctions} contains the values of the slice positions.

\begin{table*}
    \centering
    \ra{1.3}
	\begin{tabular}{@{}lcc@{} }
		\toprule
		Name & $s$ & Slice positions\\
		\midrule
		\multicolumn{2}{@{}l}{Ackley}&\\
		& 4 &  -32.77,   0.00,  10.92,  32.77\\
		& 6 & -32.77, -19.66,   0.00,   6.55,  19.66,  32.77\\
		\multicolumn{2}{@{}l}{Alpine N. 1}&\\
		& 4 & -10.00,   0.00,   3.33,  10.00\\
		& 6 & -10,  -6,   0,   2,   6,  10\\
		\multicolumn{2}{@{}l}{Deflected Corrugated Spring}&\\
		& 4 & 0.00,  5.00,  6.67, 10.00\\
		& 6 & 0,  2,  5,  6,  8, 10\\
		\multicolumn{2}{@{}l}{Double-Sum}&\\
		& 4 & -65.54,   0.00,  21.85,  65.54\\
		& 6 & -65.54, -39.32,   0.00,  13.11,  39.32,  65.54 \\
		\bottomrule
	\end{tabular}
	\caption{Names and slices of the considered test functions.}
	\label{tab:testfunctions}
\end{table*}

For each of the sliced functions, we compute pairwise empirical correlation coefficients for all pairs of two slices. This is done using a grid of $100 \times 100 $ equidistant values spanning the whole domain of the two continuous variables of the test function. Then, the function values in these points are computed for all slices in order to estimate the cross-cor\-re\-la\-tions. It turns out that all the slices of the Ackley, Alpine N. 1, and the Deflected Corrugated Spring function are positively correlated. Many of them even have a perfect positive correlation. Only the Double-Sum function also has some negative correlations between some of its slices.

A high positive or negative cross-cor\-re\-la\-tion between two slices means that -- assuming the cross-cor\-re\-la\-tion has been estimated correctly by the procedure -- information from one slice can be beneficial for the model accuracy in the other slice and vice versa.

Since many of the empirical cross-cor\-re\-la\-tions of the selected test functions are positive, and some models are expected to be better at estimating positive over negative correlations, the testbed might not be suitable for a fair comparison. For this reason, we also consider versions of the functions with inserted negative cross-\-cor\-relations by turning some of their slices upside down. This could be done by just negating all the values of the slices to be turned. Here, we introduce a more sophisticated approach that ensures that the global optimum of the original function remains the global optimum in the manipulated function. For a minimization problem, we do this as follows: The idea is to take the function value in the global optimum, $y^*$, and add a function that is strictly positive. Assume that we want to turn the $i$-th slice, $i\in\{1,\dots, s\}$, and that the global optimum is in a different slice. First, we take the function value of the original sliced function, $f(i, \cdot, \cdot)$, and subtract it from the maximum function value of the slice, $y_i^{\text{max}}$. This difference is in $\left[0, y_i^{\text{max}} - y_i^{\text{min}} \right]$. Since we do not know $y_i^{\text{max}}$, we have to estimate it by running a nonlinear optimization method, returning $\widehat{y}_i^{\text{max}}$. Then, $z(i, x_1, x_2) = \left(\widehat{y}_i^{\text{max}} - f(i, x_1, x_2)\right) \in \left[\widehat{y}_i^{\text{max}} - y_i^{\text{max}}, \widehat{y}_i^{\text{max}} - y_i^{\text{min}} \right]$. Because $\widehat{y}_i^{\text{max}} \le y_i^{\text{max}}$, $z(i, x_1, x_2)$ might take negative values. To prevent this from happening, we multiply $z(i, x_1, x_2)$ with the cumulative density function of the exponential distribution with $\lambda = 0.5$ in $z(i, x_1, x_2)$, which smoothly sets negative values to zero. Since we want the global optimum to be unique, we add a tenth of $\widehat{y}_i^{\text{max}}$. All in all, we get as the function value in the turned $i$-th slice in point $(x_1, x_2)$:
\[
y^* + z(i, x_1, x_2)  \cdot \left(1 - \exp\left(-0.5 \cdot z(i, x_1, x_2) \right)\right) + \frac{\widehat{y}_i^{\text{max}}}{10},
\]
which is in $\left[y^* + \frac{\widehat{y}_i^{\text{max}}}{10}, y^* +  \widehat{y}_i^{\text{max}} - y_i^{\text{min}} + \frac{\widehat{y}_i^{\text{max}}}{10}\right]$.

With this procedure, we generate a new function for the Ackley, Alpine N. 1, and the Deflected Corrugated Spring function. For $s = 4$ and $s = 6$, we randomly select two and three slices, respectively, making sure we do not select the slices that contain the global optimum. By turning $\frac{s}{2}$ slices of a function with only perfect positive cross-cor\-re\-la\-tions, we get another function with the highest possible amount of very high negative cross-cor\-re\-la\-tions. In fact, the turned function has $\frac{s^2}{4}$ negative and $\frac{s^2 - 2s}{4}$ positive cross-cor\-re\-la\-tions. In this case, we turned slices 1 and 3, and 1, 2, and 4, respectively. This results in 4 of 6 cross-cor\-re\-la\-tions being negative for $s=4$. For $s=6$, 9 of 15 cross-cor\-re\-la\-tions are negative.

\begin{figure*}
	\centering
	\includegraphics[width = 0.9\textwidth]{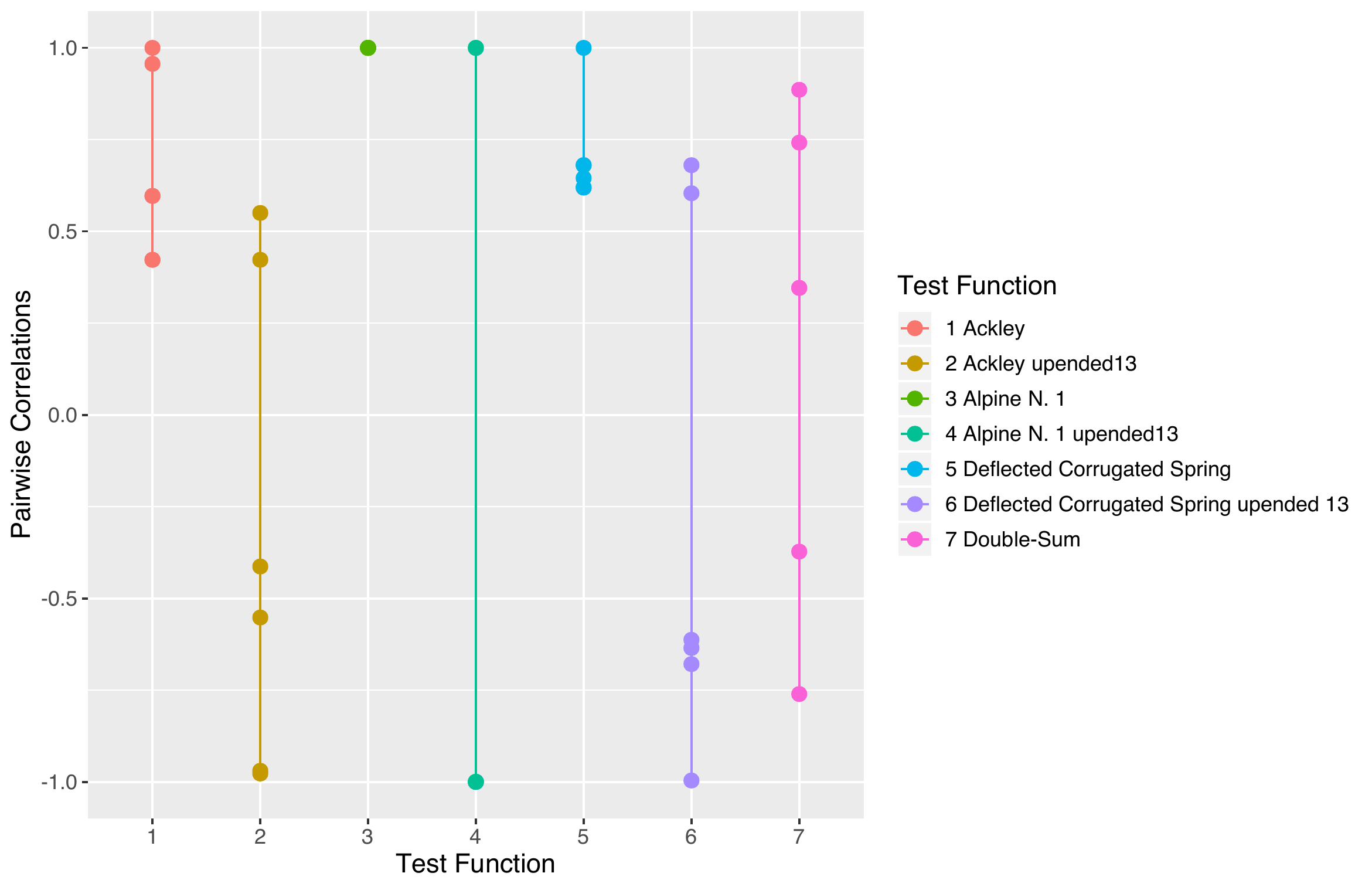}
	\caption{Empirical cross-cor\-re\-la\-tions of the test functions for $s=4$ levels of the categorical variable.}
	\label{fig:pairwisecorrs4}
\end{figure*}

\begin{figure*}
	\centering
	\includegraphics[width = 0.9\textwidth]{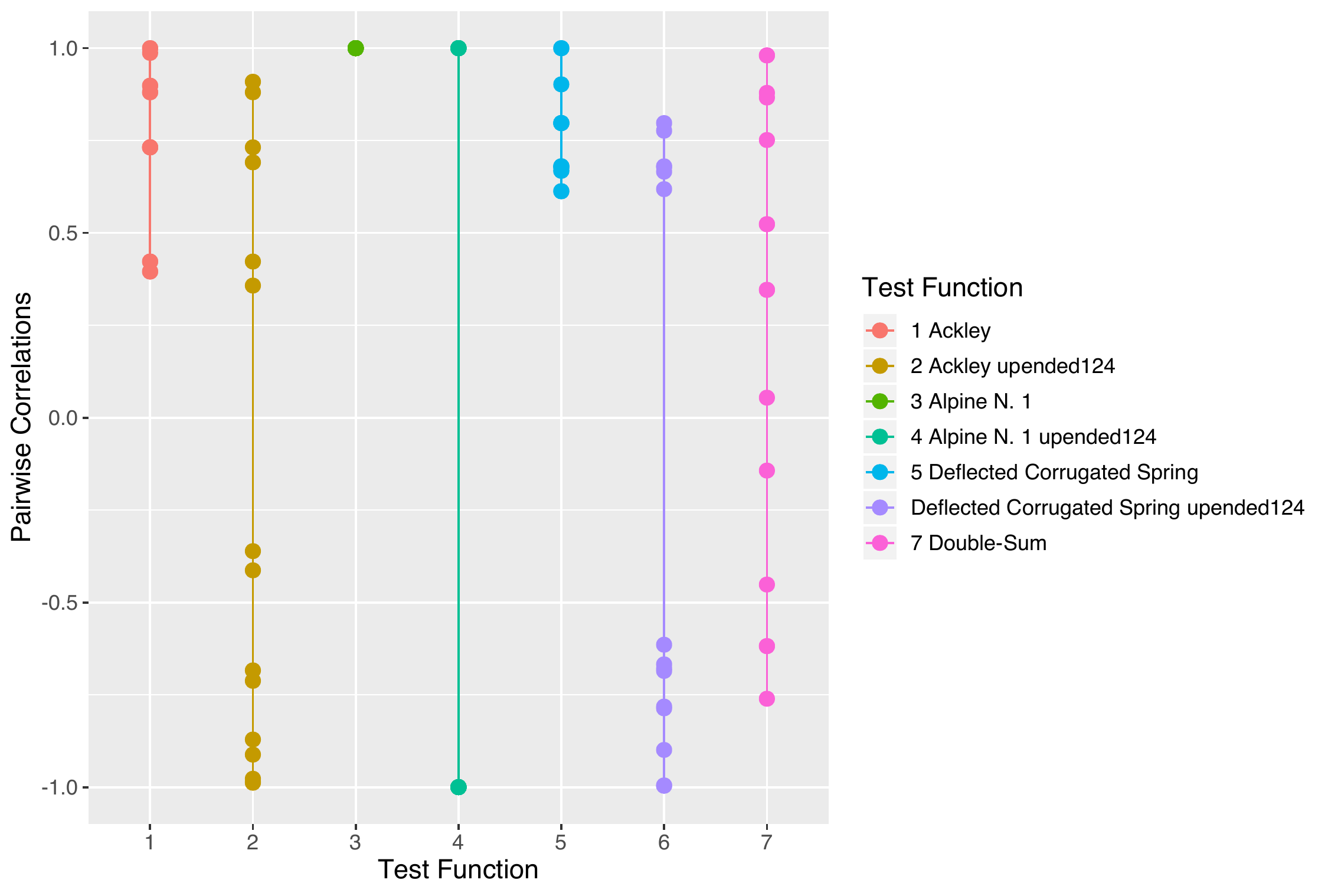}
	\caption{Empirical cross-cor\-re\-la\-tions of the test functions for $s=6$ levels of the categorical variable.}
	\label{fig:pairwisecorrs6}
\end{figure*}

Figures \ref{fig:pairwisecorrs4} and \ref{fig:pairwisecorrs6} show the empirical pairwise cross-cor\-re\-la\-tions for all considered test functions with $s=4$ and $s=6$ slices, respectively. The manipulated functions are indicated with the word ``upended'' behind the original function's name followed by the slices that were turned.

\section{Comparison between EC, MC, UC, and LRC$_r$}
\label{sec:comp}
In this section, we want to compare the performances of all models considered in this paper regarding the estimation of the cross-cor\-re\-la\-tion matrix and the prediction accuracy. We therefor apply these models to the set of test functions described in Section \ref{sec:testfuns}.

\subsection{Estimation of Cross-Correlations}\label{sec:crosscorrs}
We first compare the estimated cross-cor\-re\-la\-tion values to the pairwise empirical correlation coefficients obtained in Section \ref{sec:testfuns}. We generate 100 random CSLHDs with $n=4$ or $n=8$ points per slice, respectively. The generation of the CSLHDs has been implemented in \textsf{R} according to the algorithm given in \cite{huang}. The EC, MC, UC, and LRC$_r$ models are fitted to the resulting values for each of the sliced functions. Since for the rank $r$ of LRC$_r$ must be lower than $s$, we consider LRC$_2$ and LRC$_3$ for $s=4$ while for $s=6$ also LRC$_4$ and LRC$_5$ are considered. Table \ref{tab:numberpars2} contrasts the number of parameters that have to be estimated for the categorical variable using the different models. 

\begin{table*}
    \ra{1.3}
    \centering
	\begin{tabular}{@{}lcrr@{}}
		\toprule
		Model& &\multicolumn{2}{c@{}}{Number of parameters}\\
		\cmidrule{3-4}
		         & & $s$ = 4 & $s$ = 6\\
		\midrule
		EC            &       & 1 & 1\\
		LRC$_2$	 &		& 3 & 5\\
		MC           &       & 4 & 6\\
		LRC$_3$	 &		& 5 & 9\\
		LRC$_4$	 &		 & -- & 12\\
		LRC$_5$	 &		 & -- & 14\\
		UC           &        & 6 & 15\\
		\bottomrule
	\end{tabular}
	\caption{Numbers of parameters needed for modeling the categorical inputs.}\label{tab:numberpars2}
\end{table*}

The goodness of the estimation is measured using the RMSE between estimated cross-cor\-re\-la\-tions $\widehat{\tau}_{ij}$ and the empirical correlation coefficients $\tilde{\tau}_{ij}$:

\[
\text{RMSE}\left(\widehat{\boldsymbol{\tau}}, \tilde{\boldsymbol{\tau}}\right) = \sqrt{\sum\limits_{i=2}^s\sum\limits_{j<i}\left(\widehat{\tau}_{ij} - \tilde{\tau}_{ij}\right)^2}.
\]
Thus, a good estimation of the cross-cor\-re\-la\-tions would result in low RMSEs.

\begin{figure}[h]
	\centering
	\includegraphics[width = 0.6\columnwidth]{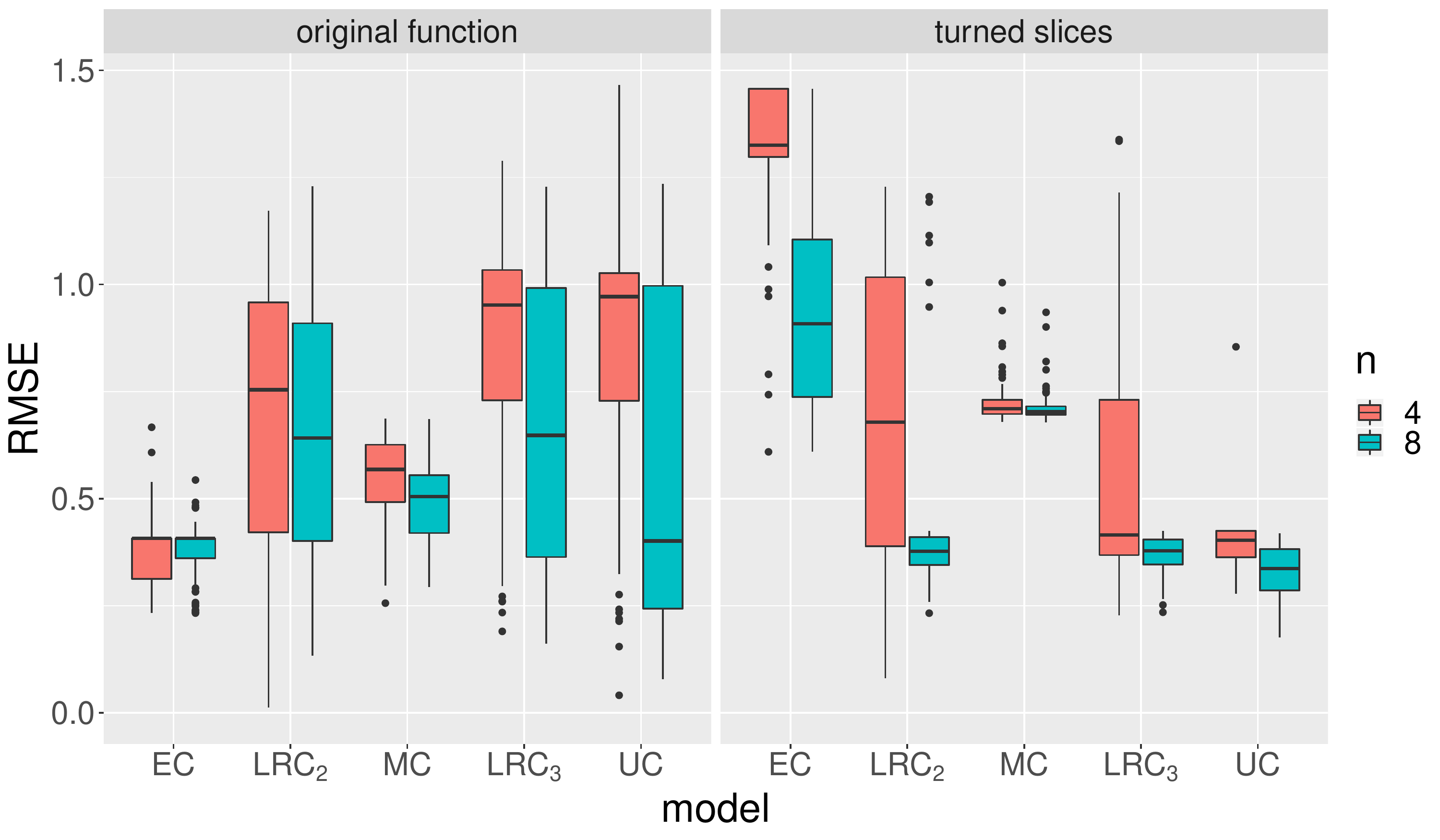}
	\caption{RMSEs of the cross-correlation estimates of the Ackley function with $s = 4$ slices}
	\label{fig:CorrRMSE_Ackley4}
\end{figure}

Figure \ref{fig:CorrRMSE_Ackley4} shows the boxplots of the RMSEs for the Ackley function with $s=4$ slices. For the original function, the cross-cor\-re\-la\-tion matrices of the EC model exhibit the smallest RMSEs followed by MC while LRC$_2$, LRC$_3$, and UC have larger errors. These models, however, improve to a much greater extent when a bigger design is used. Then, UC has the lowest median of RMSEs but still a higher variance than EC. With turned slices, LRC$_3$ and UC are distinctly better than EC and MC. LRC$_2$ has a high variance for $n=4$. For $n=8$, the variance is heavily reduced and it performs almost identically good as LRC$_3$ and even UC.

\begin{figure}[h]
	\centering
	\includegraphics[width = 0.6\columnwidth]{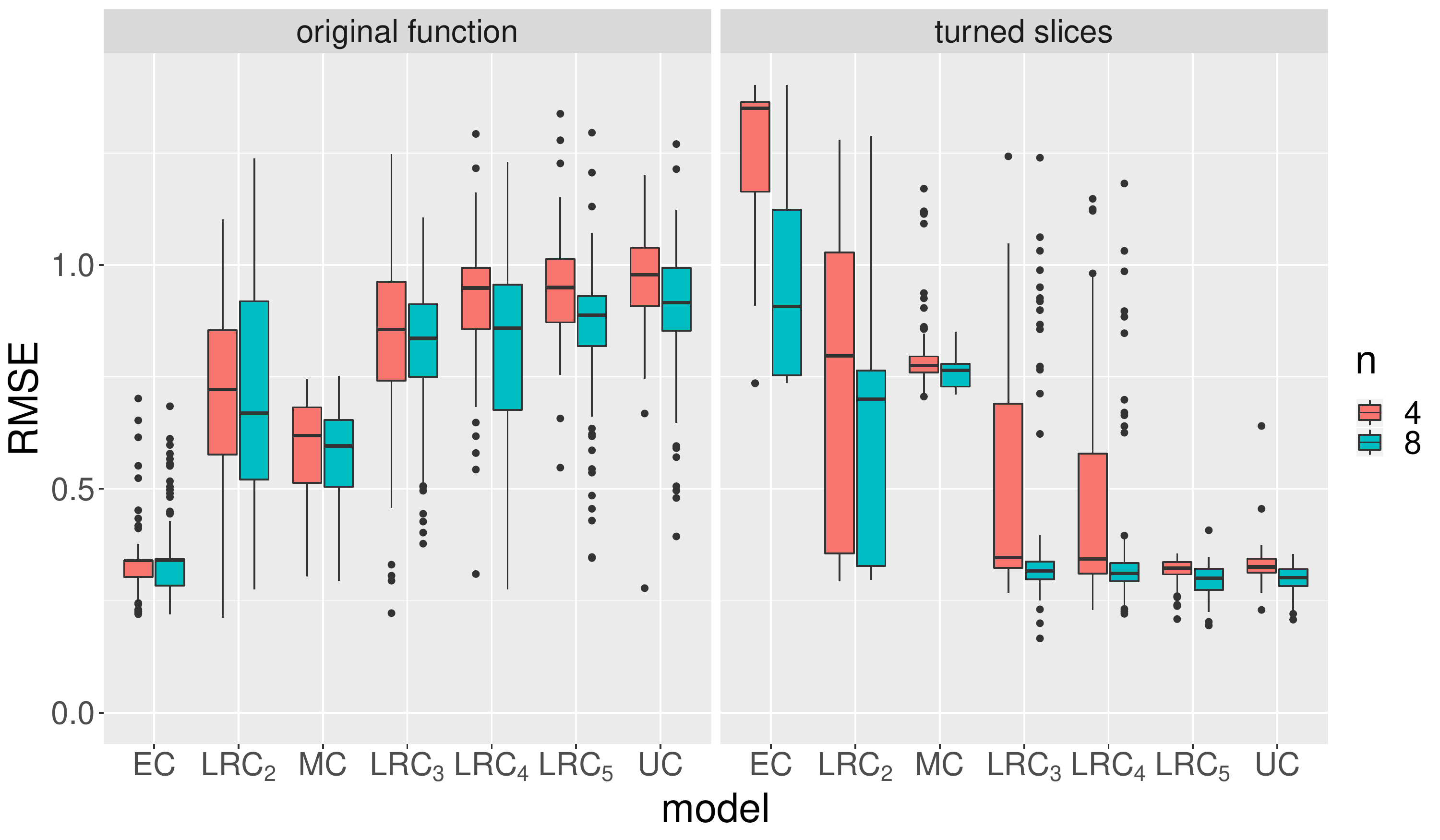}
	\caption{RMSEs of the cross-correlation estimates of the Ackley function with $s = 6$ slices}
	\label{fig:CorrRMSE_Ackley6}
\end{figure}

For $s=6$ slices, the results are similar, as can be seen in Figure \ref{fig:CorrRMSE_Ackley6}. EC performs best on the original function, followed by MC, the LRC$_r$ models with $r$ in an ascending order, and UC. Unlike with $s=4$, here the models with a higher number of parameters do not improve as much when the design size is doubled. Presumably, an even larger number of points in the design is needed for the models to show their strengths. With turned slices and $n=8$, most model fits of LRC$_3$, LRC$_4$, LRC$_5$, and UC perform similarly well. However, LRC$_3$ and LRC$_4$ have some outliers with high RMSEs and also the high variance for $n=4$ makes LRC$_5$ and UC the overall most accurate models with this function. 

\begin{figure}[h]
	\centering
	\includegraphics[width = 0.6\columnwidth]{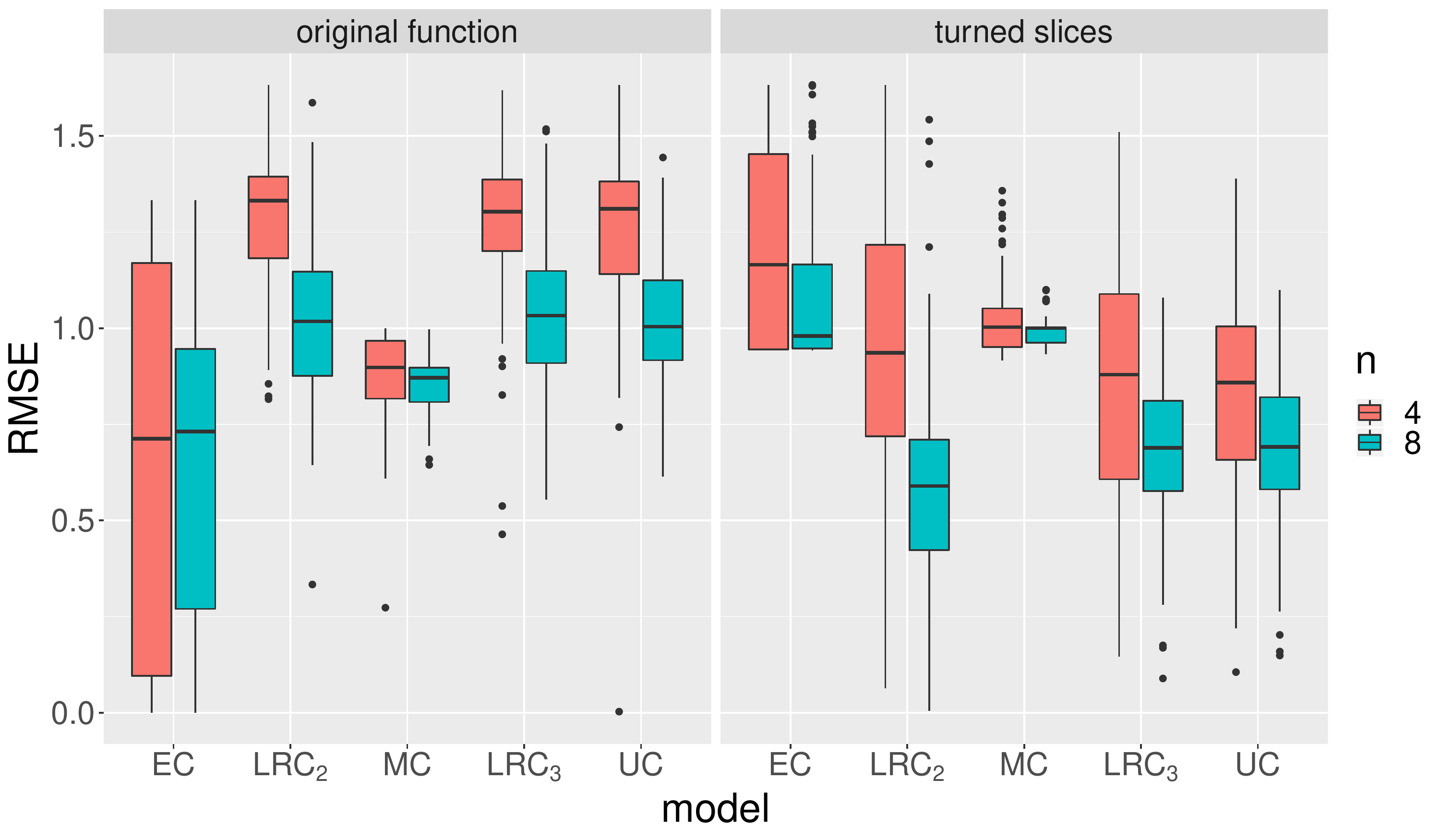}
	\caption{RMSEs of the cross-correlation estimates of the Alpine N. 1 function with $s = 4$ slices}
	\label{fig:CorrRMSE_Alpine4}
\end{figure}

\begin{figure}[h]
	\centering
	\includegraphics[width = 0.6\columnwidth]{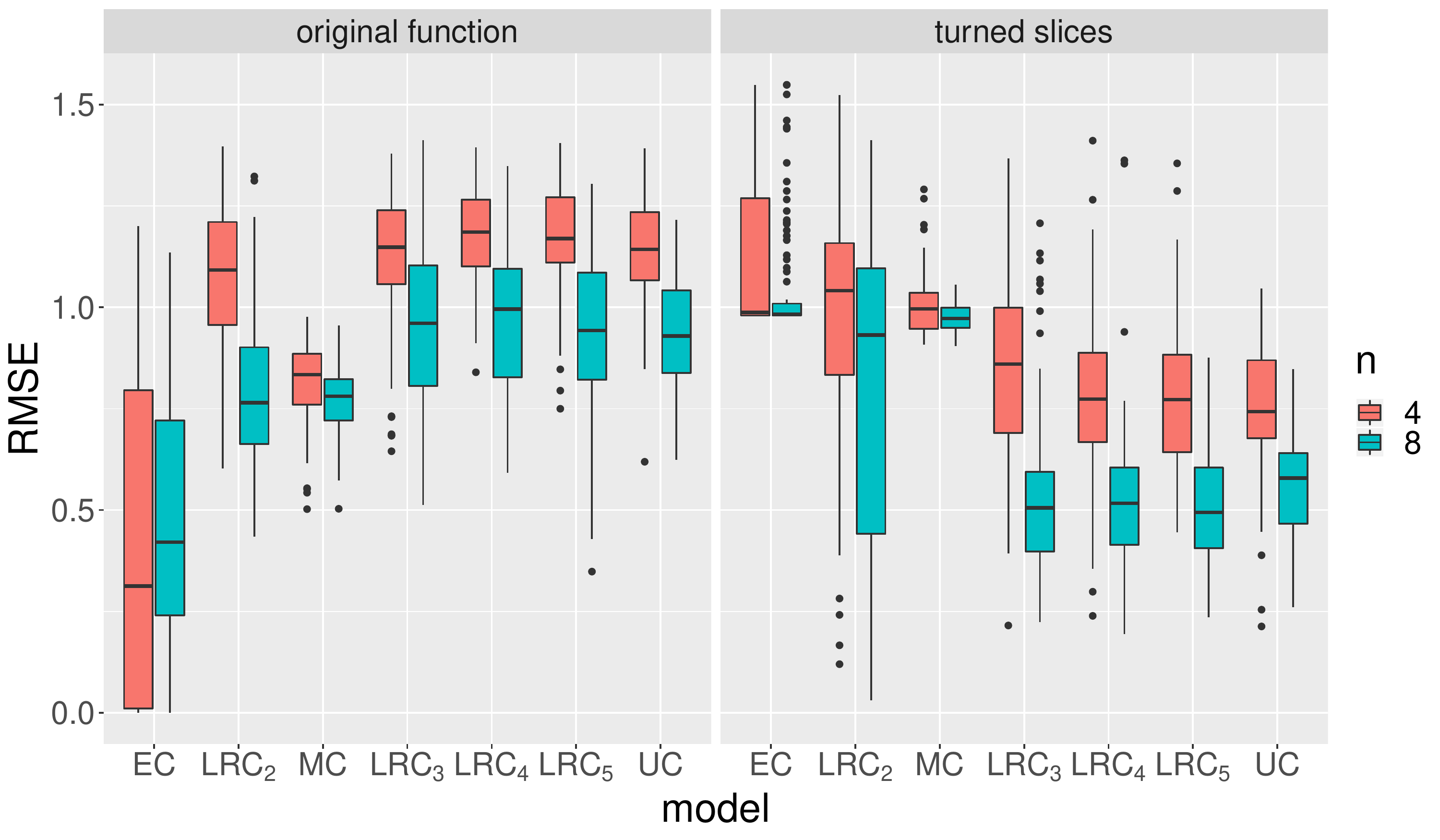}
	\caption{RMSEs of the cross-correlation estimates of the Alpine N. 1 function with $s = 6$ slices}
	\label{fig:CorrRMSE_Alpine6}
\end{figure}

The RMSEs of the Alpine N. 1 function with $s=4$ and $s=6$ are shown in Figures \ref{fig:CorrRMSE_Alpine4} and \ref{fig:CorrRMSE_Alpine6}, respectively. For $s=4$, again EC and MC performed better on the unmanipulated function. The rest of the models perform similarly well and improve greatly when $n=8$ points per slice are in the design. For the function with turned slices, EC and MC are distinctly worse than the LRC$_r$ and UC models. For $n=4$, LRC$_r$ and UC are better the higher their number of parameters is. For $n=8$, on the other hand, it is the other way around: LRC$_2$ outperforms LRC$_3$ and UC.
For $s=6$ slices, the results are very similar on the original function, with the exception of LRC$_2$, which performs comparably well as MC for $n=8$. For the manipulated function and $n=8$, LRC$_r$ with a medium rank between 3 and 5 performs the best, while UC is slightly worse and LRC$_2$ is only a little better than MC.

\begin{figure}[h]
	\centering
	\includegraphics[width = 0.6\columnwidth]{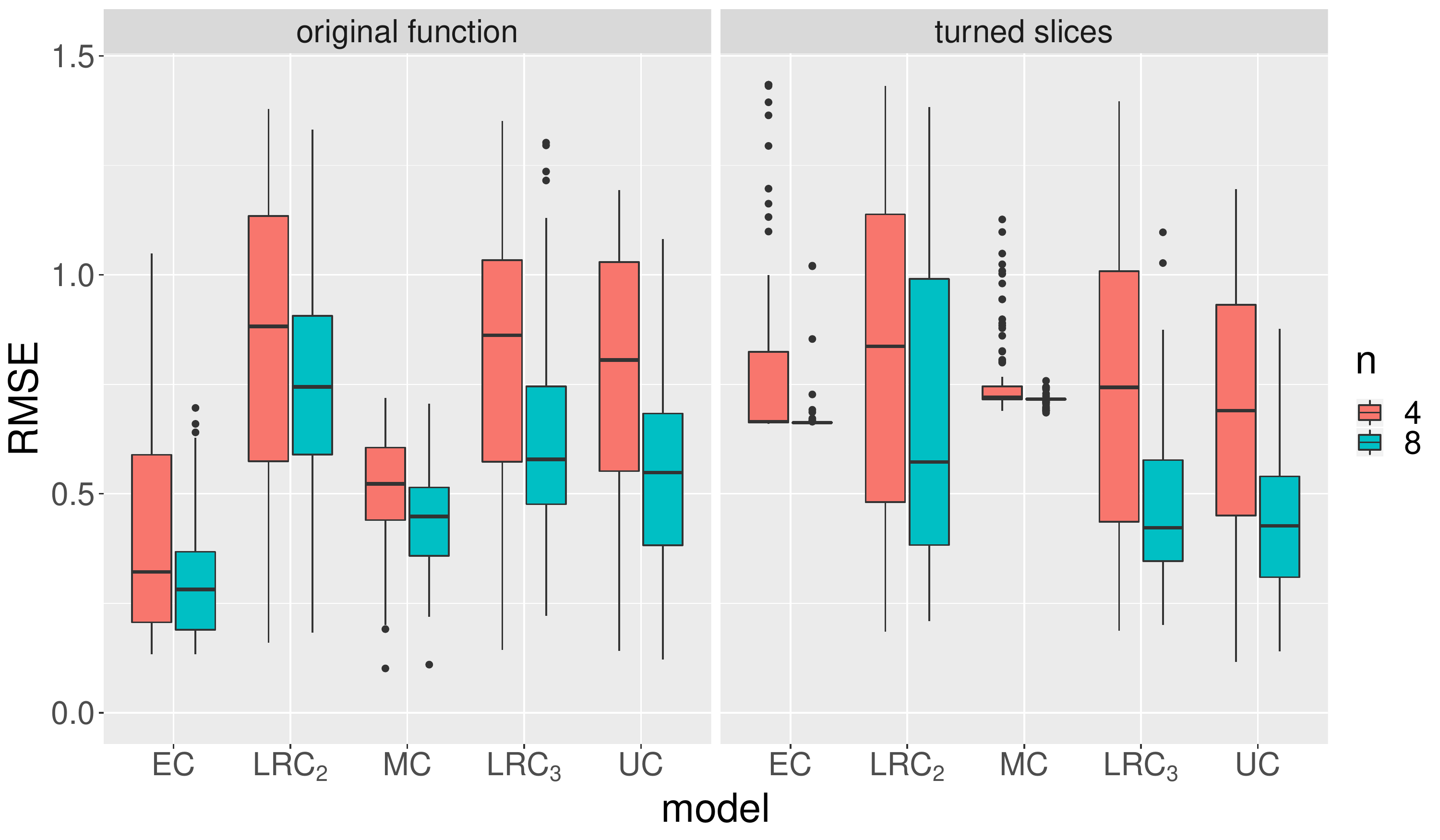}
	\caption{RMSEs of the cross-correlation estimates of the Deflected Corrugated Spring function with $s = 4$ slices}
	\label{fig:CorrRMSE_DCS4}
\end{figure}

Figure \ref{fig:CorrRMSE_DCS4} shows the results for the Deflected Corrugated Spring function with 4 slices. On the original function, EC performs best once more, independently from the size of the design. The LRC$_r$ and UC models are better the higher the number of parameters and the higher the number of design points are, but they do not get as good as MC and EC for the considered values of $n$. On the manipulated function, the LRC$_r$ and UC models this time seem to need bigger designs to display their advantage in the estimation of negative cross-cor\-re\-la\-tions as their performances are not clearly better than those of EC and MC for $n=4$. For $n=8$, however, they are. Also, the RMSEs of EC and MC are bounded at above 0.6 while the other models produce many RMSEs lower than that.

\begin{figure}[h]
	\centering
	\includegraphics[width = 0.6\columnwidth]{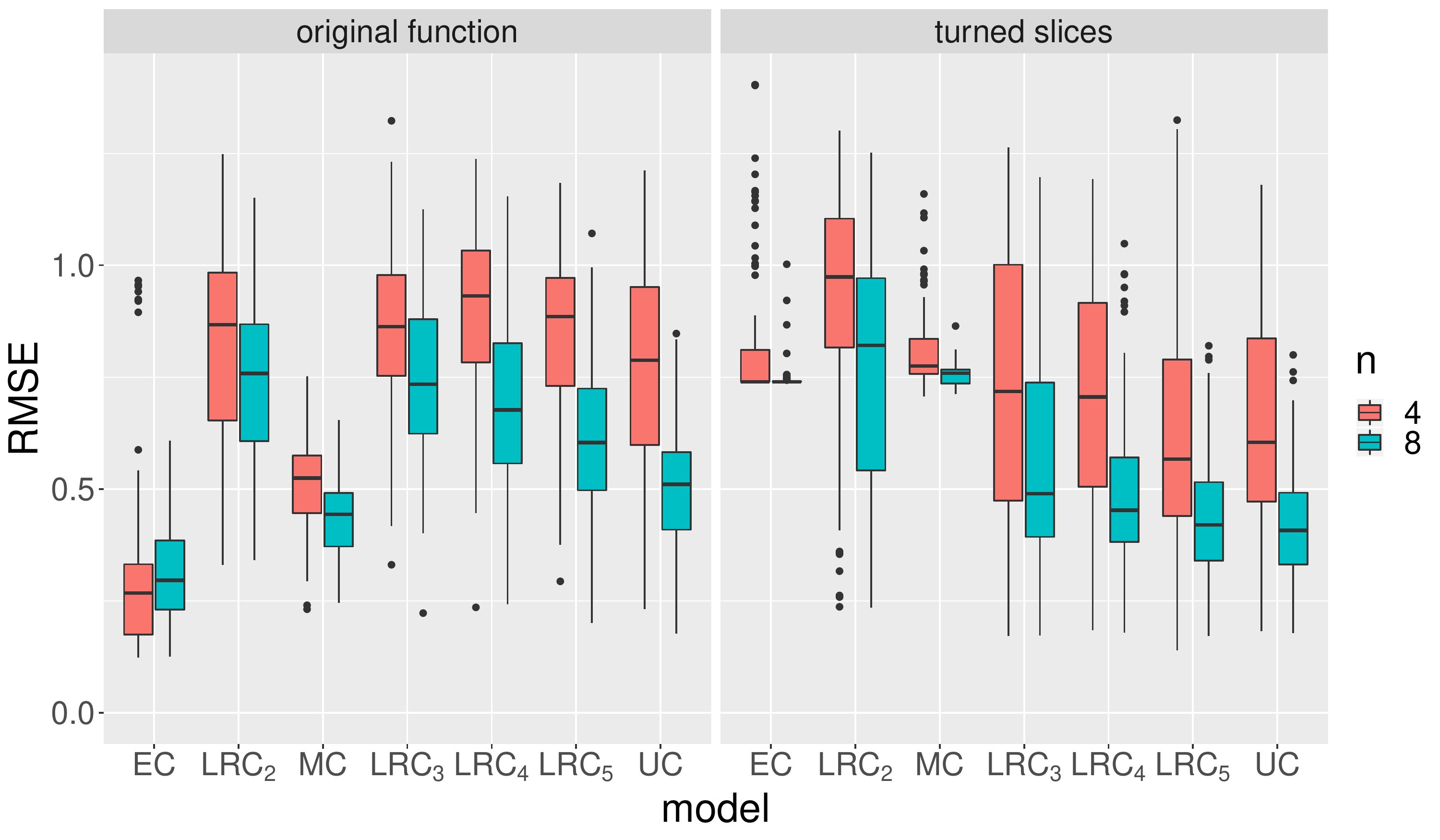}
	\caption{RMSEs of the cross-correlation estimates of the Deflected Corrugated Spring function with $s = 6$ slices}
	\label{fig:CorrRMSE_DCS6}
\end{figure}

The boxplots in Figure \ref{fig:CorrRMSE_DCS6} look very similar to those in Figure \ref{fig:CorrRMSE_DCS4}. Here, $LRC_2$ has higher RMSEs than EC and MC, even for $n=8$. The other LRC$_r$ models and UC, however, outperform EC and MC on the function with turned slices.

\begin{figure}[h]
	\centering
	\includegraphics[width = 0.6\columnwidth]{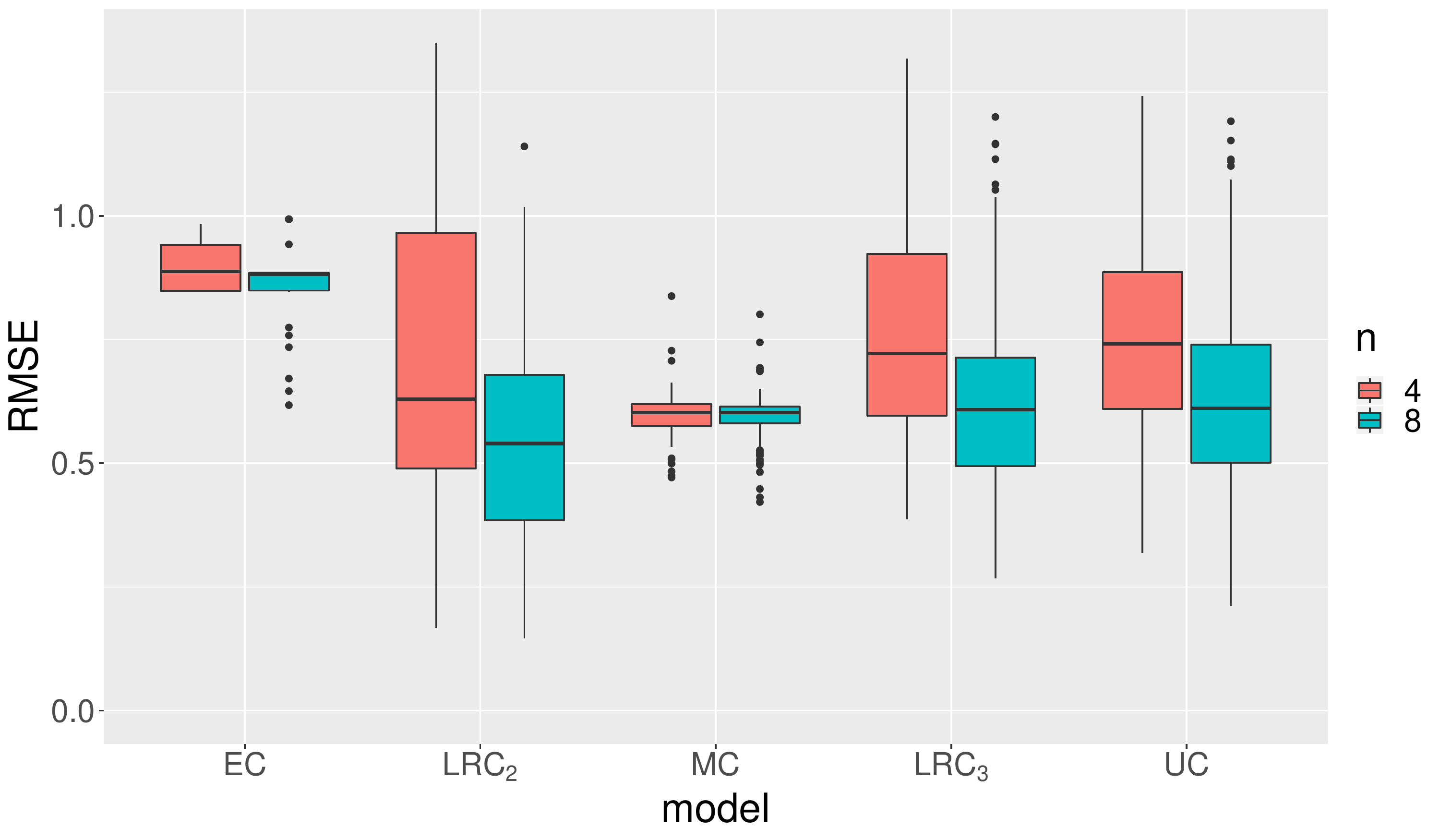}
	\caption{RMSEs of the cross-correlation estimates of the Double-Sum function with $s = 4$ slices}
	\label{fig:CorrRMSE_DS4}
\end{figure}

\begin{figure}[h]
	\centering
	\includegraphics[width = 0.6\columnwidth]{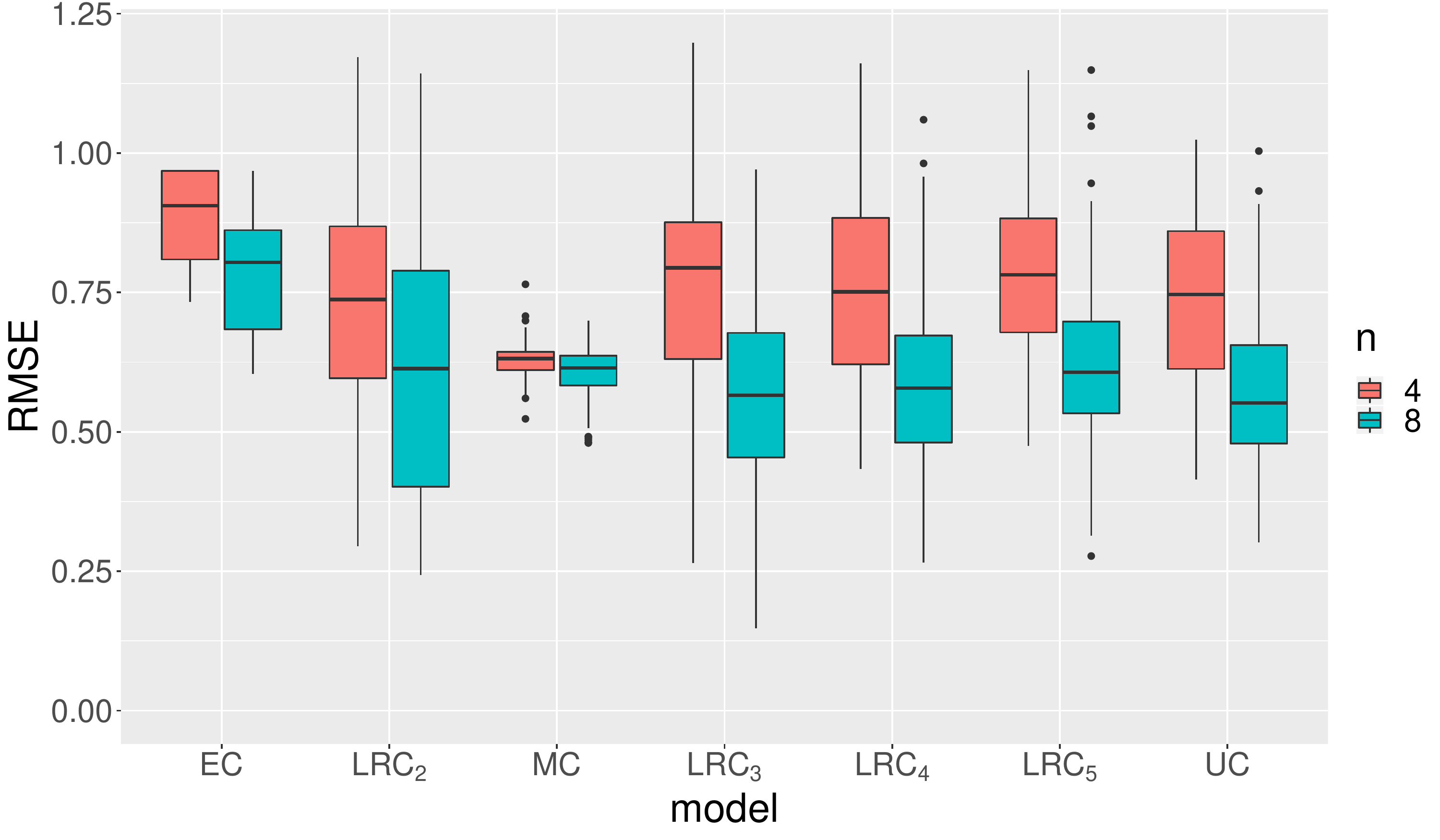}
	\caption{RMSEs of the cross-correlation estimates of the Double-Sum function with $s = 6$ slices}
	\label{fig:CorrRMSE_DS6}
\end{figure}

The results of the Double-Sum function with $s=4$ and $s=6$ are shown in Figures \ref{fig:CorrRMSE_DS4} and \ref{fig:CorrRMSE_DS6}, respectively. With $s=4$ slices and $n=4$ design points per slice, the median performance of MC is best and the variance of the RMSEs is the smallest. LRC$_r$ and UC are slightly worse, while with these models a lower number of parameters seems to result in somewhat smaller RMSEs. The EC model, however, produces the worst results. For $n=8$, LRC$_2$ performs slightly better than MC, closely followed by LRC$_3$ and UC. With $s=6$ slices, MC is also the best for $n=4$. When the design size is bigger, the LRC$_r$ and UC models improve so that LRC$_3$ through LRC$_5$ and UC perform better.

Recapitulating this subsection, the EC and MC models seem the best choice when no or not many negative cross-cor\-re\-la\-tions are assumed to be present. Else, LRC$_r$ models with a medium to high $r$ (e.g., LRC$_3$, LRC$_4$) seem to display solid results on functions with many negative cross-cor\-re\-la\-tions.

\subsection{Prediction of the Response Surface}\label{sec:rs}

In this section, we consider the accuracies of the response surfaces predicted by the different models. We therefor generate a random LHD with 1000 points. This design is -- adjusted for the bounds of the different functions' domains -- repeated for every level of the categorical variable and then used to determine the true function values.

We consider the same models, CSLHDs, and functions as in the previous section. The models' predicted values are compared to the true ones using the $Q^2$ criterion:

\[
Q^2 = 1 - \frac{\sum_{i} \left(y_i - \hat{y}_i\right)}{ \sum_{i}\left(y_i - \bar{y}\right)},
\]

where $y_i$ is the $i$-th true function value, $ \hat{y}_i$ is the model's prediction in the same point, and $\bar{y}$ is the mean of all observations on the test design. A negative value of $Q^2$ indicates that the predictions are "`worse than the mean"', i.e., we would have had a better $Q^2$ score of 0 if we just used $\bar{y}$ as the prediction at each point of the domain. The closer $Q^2$ gets to 1, the closer the predictions are to the true values.

\begin{figure}[h]
	\centering
	\includegraphics[width = 0.6\columnwidth]{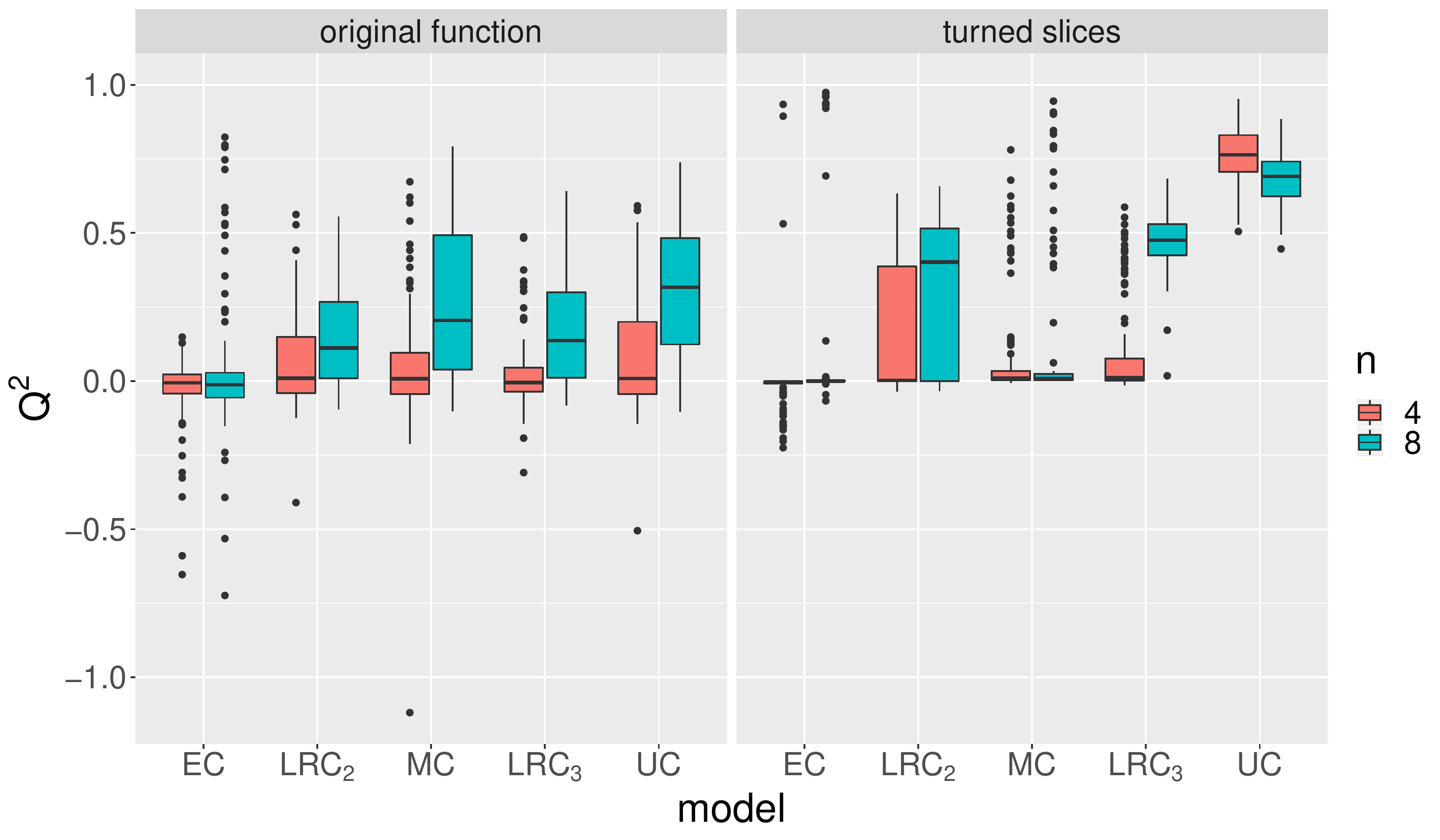}
	\caption{$Q^2$ values of the Ackley function with $s = 4$ slices}
	\label{fig:Qsq_Ackley4}
\end{figure}

Figure \ref{fig:Qsq_Ackley4} shows the values of $Q^2$ for the Ackley function with $s=4$ slices. The left hand side of the figure shows the values for the original function. For the smaller CSLHD, the medians of all functions are close to 0, i.e., no more than half of the repetitions per model resulted in a model whose predictions are better than the mean. In general, UC and LRC$_2$ followed by MC perform better than the other two models on these repetitions. For $n=8$ points per slice in the CSLHDs, all medians but the one of EC are increased. UC shows still the best results. Here, MC is slightly better than LRC$_2$ and LRC$_3$, which perform similarly well.

When two of the slices are turned such that more negative cross-cor\-re\-la\-tions are present, UC is the only model with very good results for both values of $n$. LRC$_2$ performs somewhat better than LRC$_3$ for $n=4$. For $n=8$, however, LRC$_3$ shows better results. Most repetitions of fitting EC and MC models yielded results around 0, yet there are some with (very) high values of $Q^2$.


\begin{figure}[h]
	\centering
	\includegraphics[width = 0.6\columnwidth]{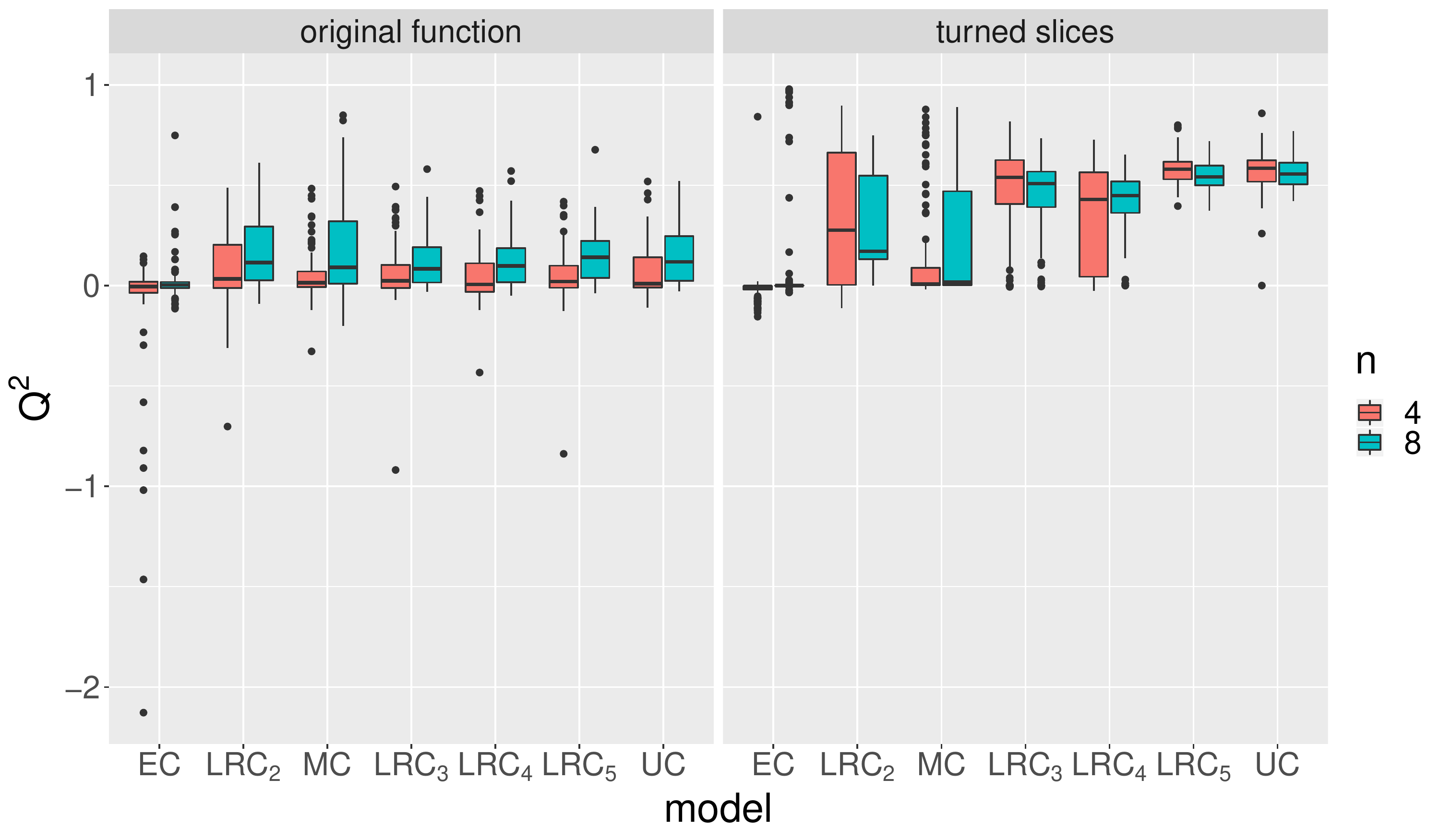}
	\caption{$Q^2$ values of the Ackley function with $s = 6$ slices}
	\label{fig:Qsq_Ackley6}
\end{figure}

Figure \ref{fig:Qsq_Ackley6} shows the results for the same function with $s=6$ slices. Here, the same seems to apply as before: For the original function, $n=4$ points per slice seem to be insufficient for accurate model fits. Yet, LRC$_2$ and UC achieve higher scores of the $Q^2$ criterion than the other models. EC performs especially bad, with some values reaching down to below -2. For $n=8$, LRC$_5$ has the highest median of $Q^2$ values but the other models (except for EC) perform similarly well. When half the slices are turned, EC and MC are worst, and UC, LRC$_5$, and LRC$_3$ are the best models. Here, a higher rank $r$ seems to improve the model accuracy of LRC$_r$ with a dip in performance at LRC$_4$. This might be because not every low-rank approximation can approximate each correlation matrix equally well -- there might be a structure in the estimated correlation matrices implied by the LRC$_4$ method that does not fit to the matrix of empirical correlations as good as the structure of, e.g., LRC$_3$.


\begin{figure}[h]
	\centering
	\includegraphics[width = 0.6\columnwidth]{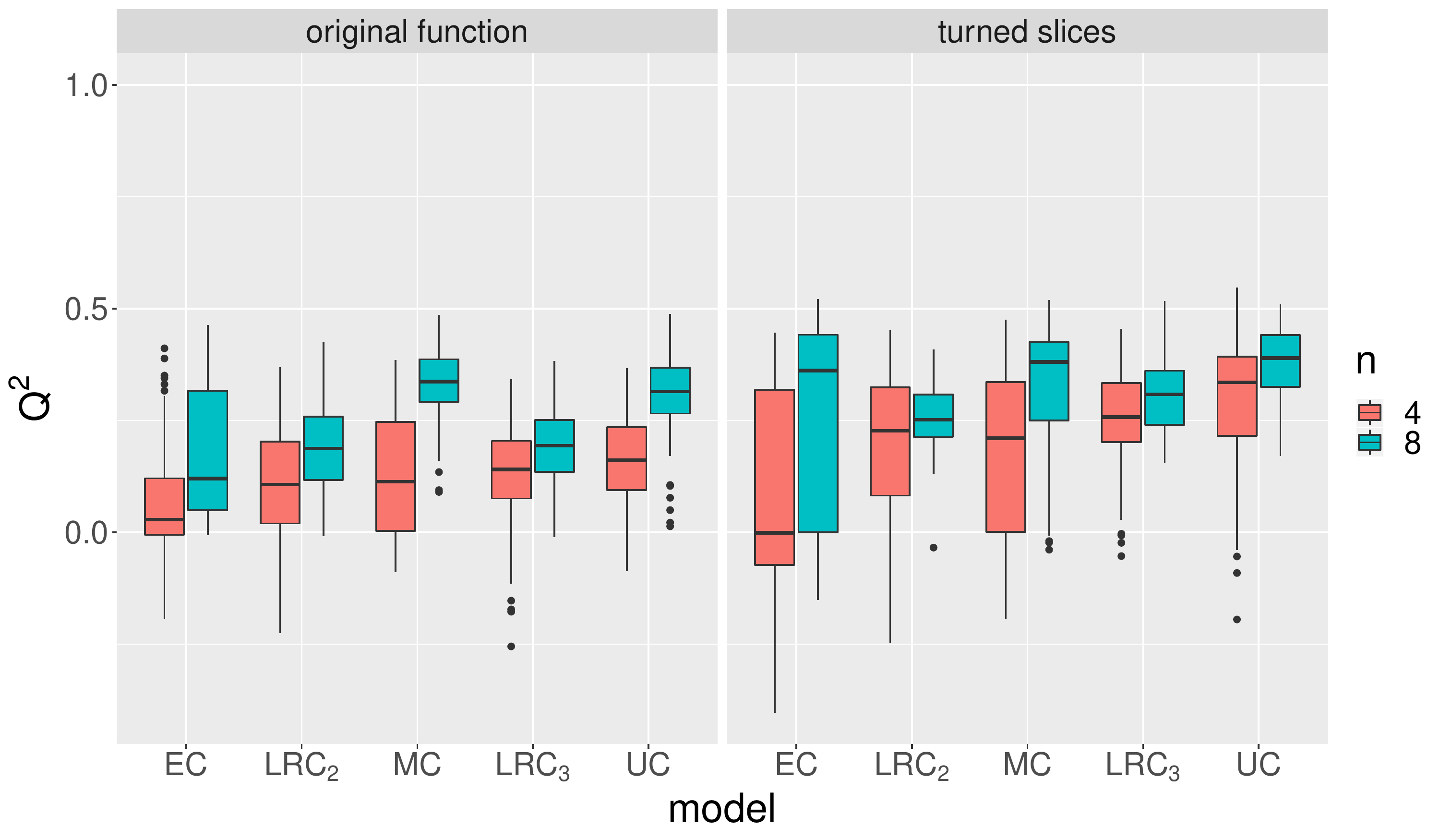}
	\caption{$Q^2$ values of the Alpine N. 1 function with $s = 4$ slices}
	\label{fig:Qsq_Alpine4}
\end{figure}

Figure \ref{fig:Qsq_Alpine4} shows the $Q^2$ values for the Alpine N. 1 function with $s=4$. For the original function and the small designs with $n=4$, the median performances are higher the higher the number of parameters is. With the bigger designs, MC shows the biggest improvement and is even better than the UC model. When
half of the function's slices are turned, the overall performance of the models gets better. UC and the LRC$_r$ models deal slightly better with fewer points than the other models. For the bigger designs, interestingly, the medians of MC and EC are close to the median of UC while LRC$_2$ and LRC$_3$ are worse.

\begin{figure}[h]
	\centering
	\includegraphics[width = 0.6\columnwidth]{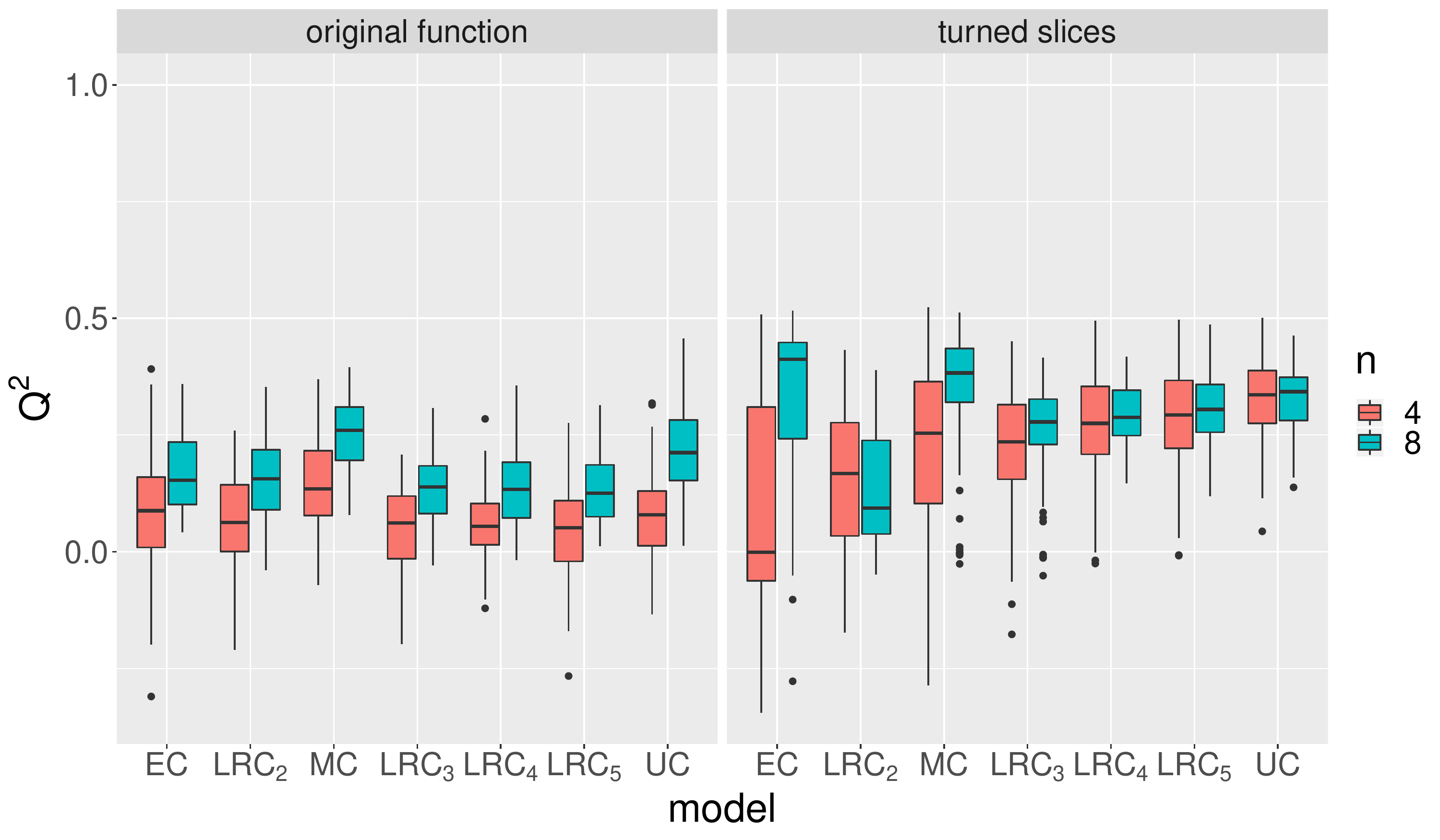}
	\caption{$Q^2$ values of the Alpine N. 1 function with $s = 6$ slices}
	\label{fig:Qsq_Alpine6}
\end{figure}

For $s=6$ slices on the unmanipulated function, MC is the best model for both values of $n$, as can be seen in Figure \ref{fig:Qsq_Alpine6}. For the function with turned slices and $n=4$, more parameters tend to lead to better results while the median performance of MC is between LRC$_3$ and LRC$_4$. For $n=8$, EC and MC improve drastically and show even better results than UC. This could not have been expected from the results of the previous section as the higher-rank LRC$_r$ and UC models show a much more accurate estimation of the correlation matrix than EC and MC (see Figure \ref{fig:CorrRMSE_Alpine6}). One explanation might be that EC and MC estimate the cross-cor\-re\-la\-tions to be 0 because they are unable to capture the correlations of -1, which in turn leads to models that ignore the points of the other levels of the categorical variable. In this case, obviously, these individual models outperform the methods that use slightly inaccurate estimates of the cross-cor\-re\-la\-tions.

\begin{figure}[h]
	\centering
	\includegraphics[width = 0.6\columnwidth]{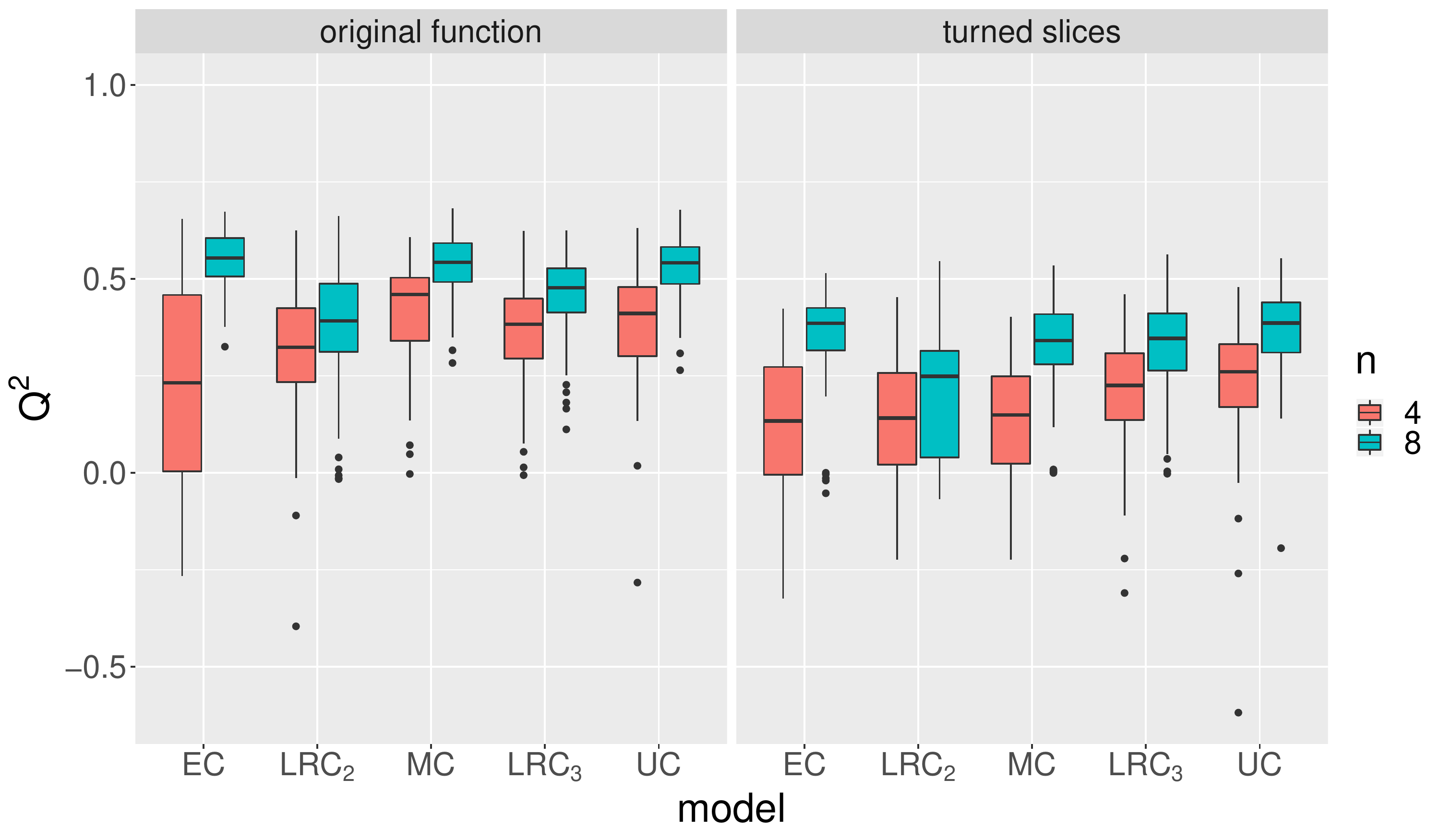}
	\caption{$Q^2$ values of the Deflected Corrugated Spring function with $s = 4$ slices}
	\label{fig:Qsq_DCS4}
\end{figure}

\begin{figure}[h]
	\centering
	\includegraphics[width = 0.6\columnwidth]{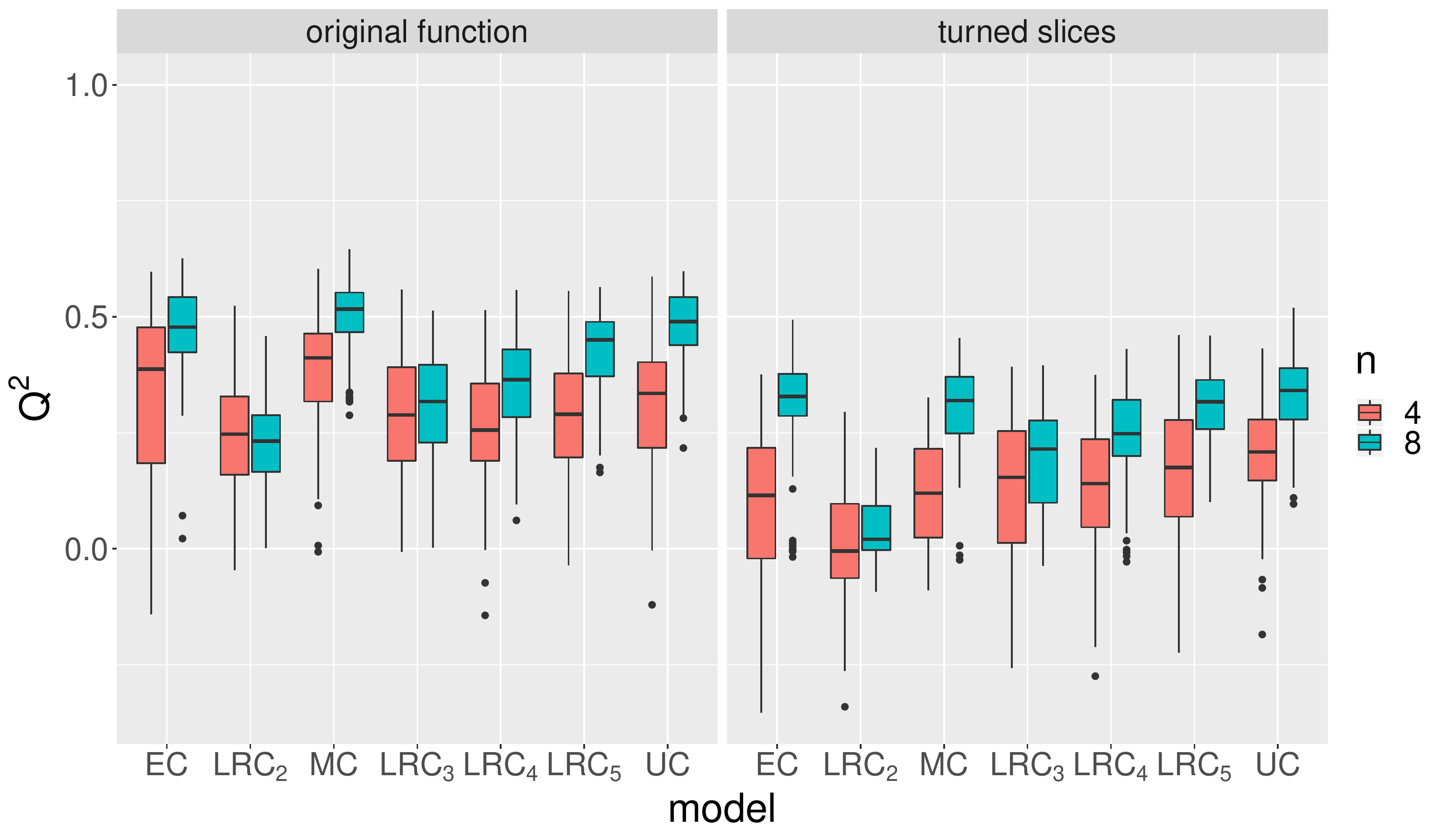}
	\caption{$Q^2$ values of the Deflected Corrugated Spring function with $s = 6$ slices}
	\label{fig:Qsq_DCS6}
\end{figure}

Figures \ref{fig:Qsq_DCS4} and \ref{fig:Qsq_DCS6} show the results for the Deflected Corrugated Spring function. On the original functions for both $s=4$ and $s=6$, MC is the best model with the smaller design size. With $n=8$, EC, MC, and UC are similarly good and better than LRC$_r$, which is better the higher the rank $r$. With turned slices, LRC$_3$ and LRC$_5$ for $s=4$ and $s=6$, respectively, UC, and MC perform comparably well. Interestingly, the values of $Q^2$ of all models are lower with the turned slices than with the original function.

\begin{figure}[h]
	\centering
	\includegraphics[width = 0.6\columnwidth]{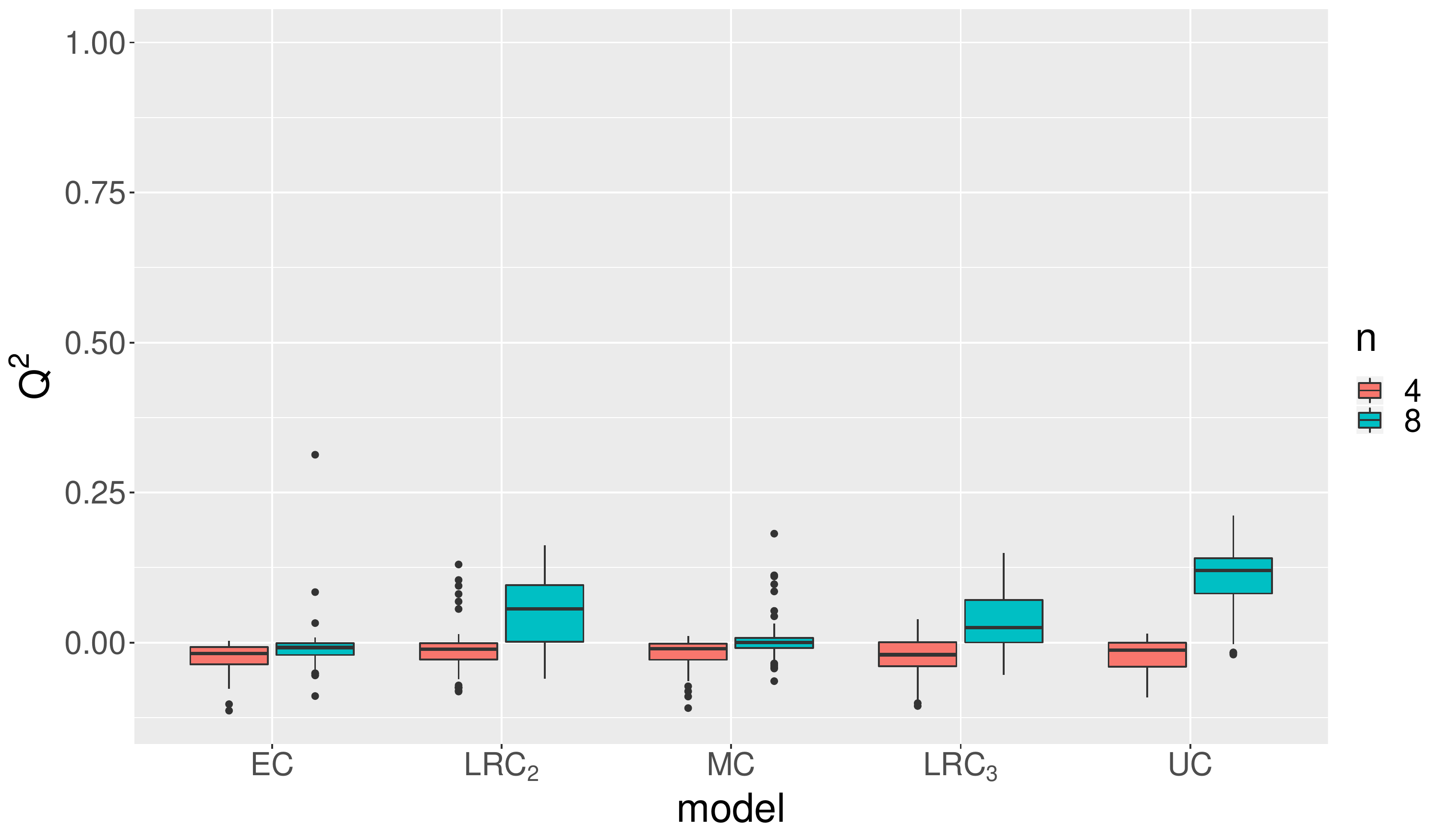}
	\caption{$Q^2$ values of the Double-Sum function with $s = 4$ slices}
	\label{fig:Qsq_DoubleSum4}
\end{figure}

The results obtained from fitting the models to the Double-Sum function are the worst among the functions considered in this paper. Figure \ref{fig:Qsq_DoubleSum4} shows the boxplots for $s=4$. For $n=4$, almost all $Q^2$ values are negative -- only LRC$_2$ and LRC$_3$ achieve some values distinctly greater than 0. For $n=8$, the results are better, and again the ones from the LRC models are better than those of MC and EC. Still, UC displays better accuracies.

\begin{figure}[h]
	\centering
	\includegraphics[width = 0.6\columnwidth]{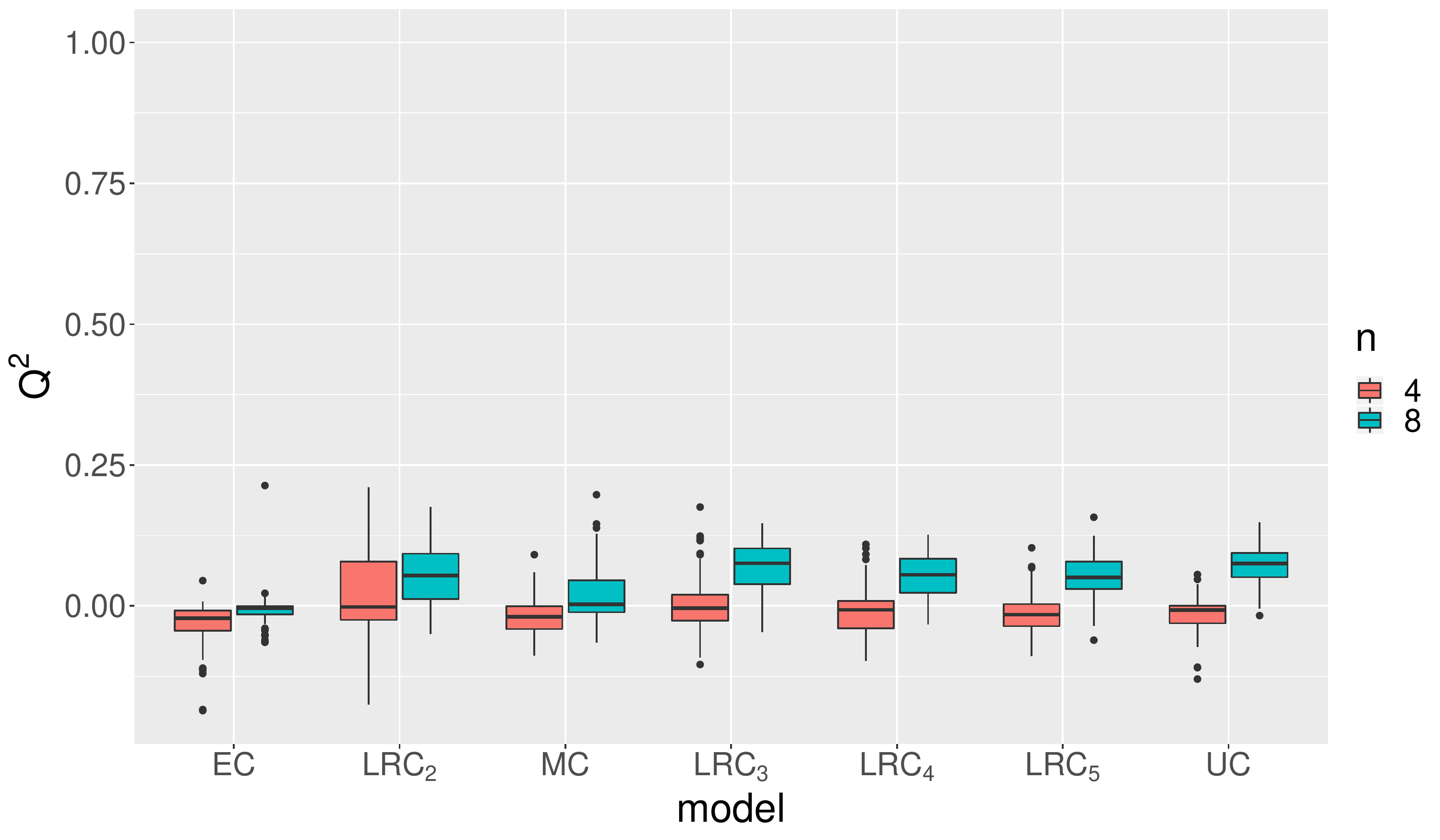}
	\caption{$Q^2$ values of the Double-Sum function with $s = 6$ slices}
	\label{fig:Qsq_DoubleSum6}
\end{figure}

Finally, Figure \ref{fig:Qsq_DoubleSum6} contains the boxplots for the Double-Sum function with $s=6$ slices. For $n=4$, LRC$_2$ shows the most positive values of $Q^2$ among the considered models. With a higher $r$, the models get weaker, while LRC$_5$ is about as good as MC. For $n=8$, LRC$_3$ and UC perform slightly better than the other LRC models. The medians of MC and EC are approximately 0.

In summary, it can be said that UC shows the overall best results, which comes at the cost of estimating a number of parameters that grows quadratically with the number of levels of the categorical variable. MC performs remarkably well despite its weak ability to estimate cross-cor\-re\-la\-tion matrices with negative elements. The LRC$_r$ models perform reasonably well most of the time and are an alternative to UC and MC especially when it is assumed that many negative cross-cor\-re\-la\-tions are present and the number of parameters of UC is not manageable. The EC model sometimes achieves surprisingly good results but in most of the cases it cannot keep up with the other models.

\section{Discussion and Outlook}\label{sec:disc}
In this paper, we introduced a new parameterization of correlation matrices into the context of Gaussian Process models with continuous and categorical inputs. The new method, LRC$_r$, produces a rank-$r$ approximation of any correlation matrix. Therefore, the number of parameters needed for the estimation of the correlation matrix is parsimonious and can be adjusted by changing rank $r$. Also, in contrast to other parsimonious methods, LRC$_r$ is not restricted to the estimation of positive correlations.

We generated a set of test functions for both continuous and categorical inputs by slicing one dimension each of a number of purely continuous test functions. This procedure can be done for any continuous test function with at least two dimensions. The global optimum can be preserved for benchmarking methods on optimization tasks. We also presented a simple way of introducing negative cross-cor\-re\-la\-tions to functions with many positive cross-cor\-re\-la\-tions.

In a comparison between the Exchangeable Correlation (EC), the Multiplicative Correlation (MC), the hypersphere decomposition-based Unrestrictive Correlation (UC), and our new method LRC$_r$, we showed that although EC and MC perform better on functions with only positive cross-cor\-re\-la\-tions, the LRC$_r$ methods -- even with only a medium rank $r$ -- have a high accuracy of the estimation of the cross-cor\-re\-la\-tion matrices with many negative elements. When considering the goodness-of-fit on a set of test points, UC performs best and MC shows remarkably good results, even when many negative cross-cor\-re\-la\-tions are present. The LRC$_r$ method performs reasonably well and is an alternative to UC and MC especially in the case of many negative cross-cor\-re\-la\-tions and when the number of parameters of UC is not manageable due to a high number of combinations of levels of the categorical inputs relative to the number of feasible design points.

It is unclear, however, how sensitive the parsimonious methods are to structural changes of the true cross-cor\-re\-la\-tion matrix to be estimated: Especially in the case of both positive and negative correlations, a different order of the levels changes the cross-cor\-re\-la\-tion matrix drastically, which might result in different performances of its estimation. Here, a higher number of parameters should lead to a greater flexibility. This should be considered in the future.

\cite{leatherman} show that designs based on the integrated mean squared prediction error may be advantageous over LHDs in certain circumstances for purely numeric variables. One could study how this behaves when extended to the mixed case.

Another possible direction is to compare how the different methods influence the performance in optimization tasks. A high goodness-of-fit not necessarily implies good results in optimization.

Moreover, a methodology on how to choose the best model in an application should be developed. This includes how to assess a model's accuracy when there are only very few points available. Even when the leave-one-out validation favors one model over the other, this not necessarily means it is the better one in an iterative optimization procedure like the Efficient Global Optimization (EGO) algorithm \citep{jones}. Maybe it is even best to change the estimation of the cross-cor\-re\-la\-tion matrix after a number of points have been added to the initial design since the methods perform differently well depending on the number of points.



\section*{Acknowledgments}
 We thank Olivier Roustant for fruitful discussions and initiating to include the implementation of LRC$_r$ into \textbf{kergp}, and Tristan Fryszman for the implementation of functions for slicing continuous test functions. The financial support of the Deutsche Forschungsgemeinschaft (DFG, German Research Foundation, 210421359/GRK1855) is gratefully acknowledged.

\appendix
\section{Computational Issues}
Here, we give some details on the implementation in \textsf{R}. Section \ref{sec:slicedSmoof} focuses on the slicing of test functions and Section \ref{sec:kergp} deals with GP models with mixed continuous and categorical inputs.

\subsection{Sliced Test Functions}\label{sec:slicedSmoof}

We built our implementation upon the \textsf{R} package \texttt{smoof} \citep{smoof}, which not only contains many different continuous optimization functions but also their box constraints and the positions and values of their global optima, amongst others.
To define a sliced function, we take a \texttt{smoof} function of class \texttt{smoof\_function}, which is an S3 object, and create a new S3 object of class \texttt{sliced\_\-smoof\_\-function}. This class inherits some important elements of the smoof function, especially its bounds and whether it is typically used for minimization or maximization benchmarking. On top of that, it also contains the $s$ slice positions \texttt{pos}.

The slice positions can be generated using an arbitrary quantile function \texttt{qdist}, e.g., the quantile function of the standard normal distribution \texttt{qnorm}, and proceed as follows:

\begin{verbatim}
pos = qdist(seq(0, 1, length.out = s + 2))
pos = pos[2:(s + 1)]
pos = (pos - pos[1])/(pos[s] - pos[1])
pos = pos * (max - min) + min
\end{verbatim}

where \texttt{min} and \texttt{max} are the lower and the upper bound of the dimension to be sliced, respectively.

In the first line, the quantile function is used in order to get $s$ + 2 values corresponding to the 0\%, $\frac{1}{s-1}\cdot 100\%$, $\frac{2}{s-1}\cdot 100\%$, $\dots$, and 100\%  quantile. Since the first and the last value might be $-\infty$ and $+\infty$, respectively, we omit them in the second line of the code chunk so \texttt{pos} consists only of the middle $s$ values. In the third line, these values are normalized to the interval $\left[0,1\right]$ and in the fourth line transformed to $\left[\texttt{min},\texttt{max}\right]$. With \texttt{qdist=qunif}, i.e., the quantile function of the continuous uniform distribution, the slice positions in \texttt{pos} are equidistant.

This procedure, however, ignores the position of the global optimum. Since it can be crucial to know the exact position and value of the global optimum for benchmarking purposes, one can exchange one of the slice positions in \texttt{pos} with the position of the optimum, thus ensuring that the optimum of the continuous function still exists in the sliced version of it. Here, we exchange the slice position that is closest to the global optimum, considering only the dimension to be sliced.

\begin{verbatim}
swap.ind = which.min(abs(pos - global.opt.pos))
pos[swap.ind] = global.opt.pos
\end{verbatim}

Here, \texttt{global.opt.pos} is the position of the global optimum in the dimension to be sliced.

\subsection{GP Models with Mixed Inputs}\label{sec:kergp}

The Gaussian Process Laboratory \texttt{kergp} \citep{kergp} is a very useful \textsf{R} package for fitting Kriging models. It lets the user define their own kernels and also provides some predefined kernels. For mixed continuous and categorical inputs, it is possible to define compound kernels like in Equation \eqref{eq:compcov}. For example, a compound kernel of Mat\'{e}rn and LRC$_2$ kernels for two continuous variables and one categorical variable with $s = 4$ levels is generated as follows:

\begin{verbatim}
library(kergp)

mat.kernel = kMatern(d = 2, nu = "5/2")
inputNames(k) = c("x2", "x3")

lrc.2.kernel = q1LowRank(
  factor(levels = 1:4),
  input = "x1",
  rank = 2,
  cov = "corr"
)

comp.kernel.mat.lrc.2 = covComp(
  formula = ~ mat.kernel() * lrc.2.kernel()
)
\end{verbatim}

The LRC$_r$ kernels have already been implemented in \texttt{kergp}. For MC, however, we have to define our own kernel. For categorical kernels, this is done by implementing three functions: \texttt{q1Multi}, which is the one called for constructing the kernel along with the correct number of parameters and their bounds; \texttt{corLevMulti}, which performs a few type checks and calls the main function, and \texttt{.corLevMulti}, which contains the code that generates the MC cross-cor\-re\-la\-tion matrix given the parameter vector \texttt{par} using the following three lines of code:

\begin{verbatim}
vec = exp(-par)
R = vec %*% t(vec)
diag(R) = 1
\end{verbatim}

After that, the compound kernel is generated just like the LRC$_2$ one above:

\begin{verbatim}
mc.kernel = q1Multi(
  factor(levels = 1:4),
  input = "x1",
  cov = "corr"
)
\end{verbatim}
\begin{verbatim}
comp.kernel.mat.mc = covComp(
  formula = ~ mat.kernel() * mc.kernel()
)
\end{verbatim}


\bibliographystyle{abbrvnat}
\bibliography{bib}

\end{document}